\pdfoutput=1

\documentclass[11pt]{article}

\usepackage{acl}

\usepackage{url}
\usepackage{makecell}
\usepackage{tabularx}
\usepackage[pdftex]{graphicx}
\usepackage[cmex10]{amsmath}
\usepackage{xcolor}
\usepackage{ulem}
\usepackage{caption}
\usepackage{subcaption}
\usepackage{placeins}
\usepackage{appendix}
\usepackage{multirow}
\usepackage{dingbat}
\usepackage{booktabs}
\usepackage{times}
\usepackage{latexsym}
\usepackage{enumitem}
\usepackage{balance}
\usepackage{float}

\newcommand{\ra}[1]{\renewcommand{\arraystretch}{#1}} 

\usepackage[T1]{fontenc}

\usepackage[utf8]{inputenc}

\usepackage{microtype}

\newcommand{\darkbert}{DarkBERT} 

\title{DarkBERT: A Language Model for the Dark Side of the Internet}

\author{
    Youngjin Jin\textsuperscript{1} \hspace{0.5em}
    Eugene Jang\textsuperscript{2} \hspace{0.5em}
    Jian Cui\textsuperscript{2} \hspace{0.5em}
    Jin-Woo Chung\textsuperscript{2} \hspace{0.5em}
    Yongjae Lee\textsuperscript{2} \hspace{0.5em} 
    Seungwon Shin\textsuperscript{1} \\ \\ 
    \textsuperscript{1}KAIST, Daejeon, South Korea \\
    \textsuperscript{2}S2W Inc., Seongnam, South Korea \\
    \textsuperscript{1}\texttt{\{ijinjin,claude\}@kaist.ac.kr} \\
    \textsuperscript{2}\texttt{\{genesith,geeoon19,jwchung,lee\}@s2w.inc}
}

\begin{document}
\maketitle


\begin{abstract}
  Recent research has suggested that there are clear differences in the language used in the Dark Web compared to that of the Surface Web. As studies on the Dark Web commonly require textual analysis of the domain, language models specific to the Dark Web may provide valuable insights to researchers. In this work, we introduce {\darkbert}, a language model pretrained on Dark Web data. We describe the steps taken to filter and compile the text data used to train {\darkbert} to combat the extreme lexical and structural diversity of the Dark Web that may be detrimental to building a proper representation of the domain. We evaluate {\darkbert} and its vanilla counterpart along with other widely used language models to validate the benefits that a Dark Web domain specific model offers in various use cases. Our evaluations show that {\darkbert} outperforms current language models and may serve as a valuable resource for future research on the Dark Web.
\end{abstract}

\section{Introduction}

The \textit{Dark Web} is a subset of the Internet that is not indexed by web search engines such as Google \textit{and} is inaccessible through a standard web browser. To access the Dark Web, specialized overlay network applications such as Tor (The Onion Router)~\cite{tor-269582} are required. Tor also hosts \textit{hidden services} (\textit{onion services}) --- web services in which the client and the server IP addresses are hidden from each other~\cite{biryukov-torhiddenservice}.

This sense of identity obscurity provided to the Dark Web users comes with a catch; many of the underground activities prevalent in the Dark Web are immoral/illegal in nature, ranging from content hosting such as data leaks to drug sales~\cite{al-nabki-etal-2017-classifying, coda_naacl2022}. As such, the popularity of the Dark Web as a platform of choice for malicious activities has garnered interest from researchers and security experts alike.



To handle the ever-changing landscape of modern cyber threats, cybersecurity experts and researchers have started to employ natural language processing (NLP) methods. Gaining evidence-based knowledge such as indicators of compromise (IOC) to mitigate emerging threats is an integral part of modern cybersecurity known as \textit{cyber threat intelligence} (CTI)~\cite{10.1145/acing_ioc, bromiley2016threat}, and modern NLP tools have become an indispensable part of CTI research. As such, the use of NLP techniques has also been extended to the Dark Web~\cite{coda_naacl2022, doppelgangers-www2019, choshen-etal-2019-language, al-nabki-etal-2017-classifying, ALNABKI2019212, thievescant}. The continued exploitation of the Dark Web as a platform of cybercrime makes it a valuable and necessary domain for CTI research.

\begin{figure*}[ht]
    \centering
    \includegraphics[width=\textwidth]{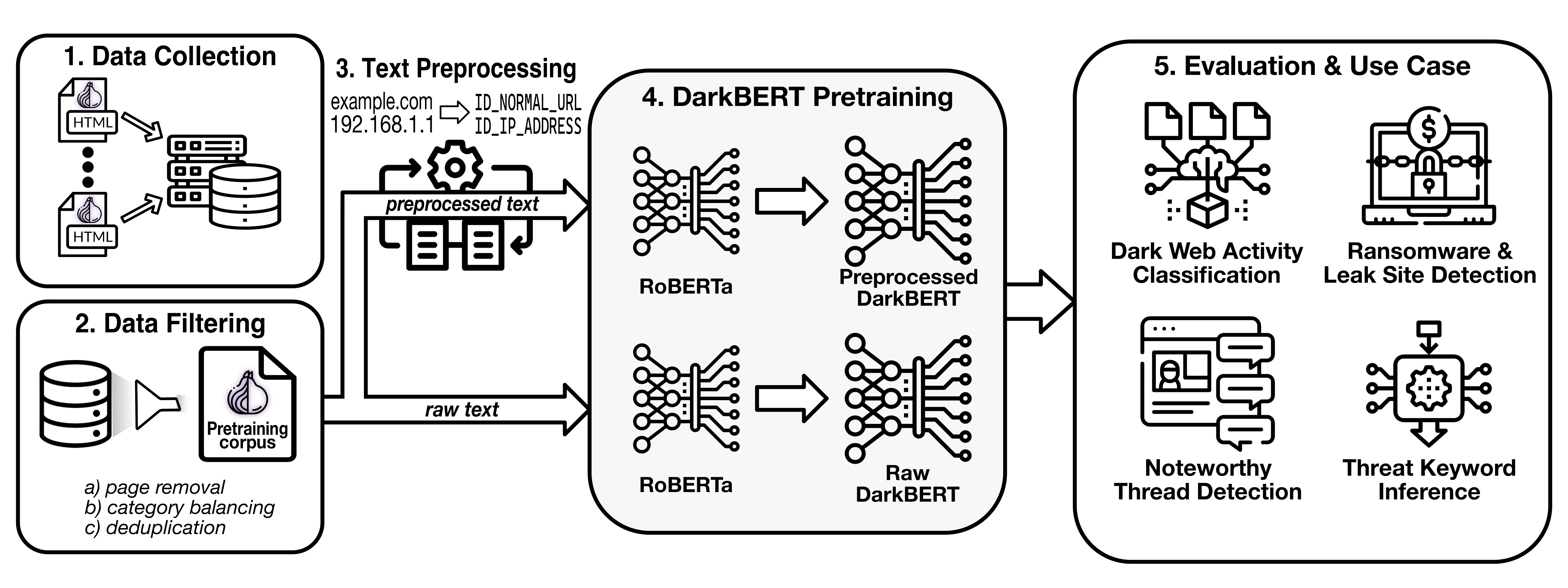}
    \caption{Illustration of the DarkBERT pretraining process and the various use case scenarios for evaluation.}
    \label{fig:darkbert}
\end{figure*}

Recently,~\citet{coda_naacl2022} observed that using a BERT-based classification model achieves state-of-the-art performance among available NLP methods in the Dark Web. However, BERT is trained on \textit{Surface Web}~\footnote{Web services and content that are readily available and indexed in common search engines such as Google} content (i.e., Wikipedia and BookCorpus)~\cite{devlin-etal-2019-bert}, which has different linguistic characteristics from that of the Dark Web~\cite{choshen-etal-2019-language}. In the context of CTI, this implies that popular pretrained language models such as BERT are not ideal for Dark Web research in terms of extracting useful information due to the differences in the language used in the two domains. Consequently, an NLP tool that is suitable for application in Dark Web domain tasks would prove to be valuable in the ongoing efforts of Dark Web cybersecurity.

In this paper, we propose \textbf{DarkBERT}, a new language model pretrained on a Dark Web corpus. To measure the usefulness of {\darkbert} in handling cyber threats in the Dark Web, we evaluate {\darkbert} in tasks related to detecting underground activities. We compare {\darkbert} to other widely used pretrained language models BERT~\cite{devlin-etal-2019-bert} and RoBERTa~\cite{roberta-paper} that are trained on data found in the Surface Web to verify the efficacy of {\darkbert} in Dark Web domain texts. Our evaluation results show that {\darkbert}-based classification model outperforms that of known pretrained language models. Furthermore, we present potential use cases to illustrate the benefits of utilizing {\darkbert} in cybersecurity-related tasks such as Dark Web forum thread detection and ransomware leak site detection.

Our contributions are summarized as follows:
\begin{itemize}[topsep=0pt,itemsep=-1ex,partopsep=1ex,parsep=1ex]
    \item We introduce {\darkbert}, a language model pretrained on the Dark Web which is capable of representing the language used in the domain compared to that of the Surface Web. 
    \item We illustrate the effectiveness of {\darkbert} in the Dark Web domain. Our evaluations show that {\darkbert} is better suited for NLP tasks on Dark Web specific texts compared to other pretrained language models.
    \item We demonstrate potential use case scenarios for {\darkbert} and show that it is better-suited for tasks related to cybersecurity compared to other pretrained language models.
    \item We provide new datasets used for our Dark Web domain use case evaluation.
\end{itemize}
\section{Related Work}
\label{sec:related_work}
The recent availability of Dark Web resources~\cite{coda_naacl2022, al-nabki-etal-2017-classifying, ALNABKI2019212} has made it possible to explore the differences between the languages used in the Dark Web and the Surface Web. \citet{choshen-etal-2019-language} explored the differences in the illegal and legal pages in the Dark Web and found a number of distinguishing features between the two domains such as named entity, vocabulary, and syntactic structure. Their analyses using standard NLP tools have also suggested that processing text in the Dark Web domain would require considerable domain adaptation. The linguistic differences between the Surface Web and the Dark Web were further examined by~\citet{coda_naacl2022} through linguistic features such as part-of-speech (POS) distribution and vocabulary usage between the texts in the two domains. 


Recently,~\citet{ranaldi-darkside} explored the use of pretrained language models over Dark Web texts to examine the effectiveness of such models, and suggested that lexical and syntactic models such as GloVe~\cite{pennington-etal-2014-glove} outperform pretrained models in some specific Dark Web tasks. Meanwhile,~\citet{coda_naacl2022} demonstrated that pretrained language models in some Dark Web tasks such as Dark Web activity classification perform better than simple lexical models, suggesting that language models like BERT show promising results in the Dark Web. Either way, a domain-specific pretrained language model would be beneficial in that it would be able to represent the language used in the Dark Web, which may effectively reduce the performance issues faced in previous experiments.
\section{DarkBERT Construction}
\label{sec:methodology}

In this section, we describe the process for building our Dark Web domain-specific pretrained language model, \textbf{DarkBERT}. We begin by collecting pages to build the text corpus used for pretraining {\darkbert} (Section~\ref{sec:methodology-data-collection}). Then, we filter the raw text corpus and employ text preprocessing methods for pretraining purposes (Section~\ref{sec:methodology-dataprocess}). Finally, we pretrain {\darkbert} using the text corpus (Section~\ref{sec:methodology-pretraining}). An overview of the {\darkbert} construction process is illustrated in Figure~\ref{fig:darkbert}.

\subsection{Data Collection}
\label{sec:methodology-data-collection}
A massive text corpus consisting of pages from the Dark Web is necessary for pretraining {\darkbert}. We initially collect seed addresses from Ahmia~\footnote{\url{https://ahmia.fi/}} and public repositories containing lists of onion domains. We then crawl the Dark Web for pages from the initial seed addresses and expand our list of domains, parsing each newly collected page with the HTML title and body elements of each page saved as a text file. We also classify each page by its primary language using fastText~\cite{joulin2016fasttext, joulin2016bag} and select pages labeled as English. This allows {\darkbert} to be trained on English texts as the vast majority of Dark Web content is in English~\cite{coda_naacl2022, 10.1145/3322645.3322691}. A total of around 6.1 million pages was collected. The full statistics of the crawled Dark Web data is shown in Table~\ref{tab:darkweb-data-stats} of the Appendix.

\subsection{Data Filtering and Text Processing}
\label{sec:methodology-dataprocess}
While the text data collected in Section~\ref{sec:methodology-data-collection} is of considerable size, a portion of the data contains no meaningful information such as error messages or duplicates of other pages. Therefore, we take three measures --- \textit{removal of pages with low information density}, \textit{category balancing}, and \textit{deduplication} --- to retain useful page samples in the pretraining corpus and remove unnecessary pages. In addition, it is critical that the model does not learn representations from sensitive information. Although a previous study stated that language models pretrained with sensitive data are unable to extract sensitive information with simple methods, the possibility cannot be ruled out using more sophisticated attacks~\cite{lehman-etal-2021-bert}. To this end, we preprocess the pretraining corpus to address ethical considerations using identifier masks or removing texts entirely, depending on the type of the target text. The details of filtering and text preprocessing are described in Sections~\ref{sec:app-datafiltering} and~\ref{sec:app-idmask} of the Appendix.




\begin{table}
\centering
\caption{The two variations of Dark Web text corpus used to train {\darkbert}.}
\label{tab:darkbert-pretrain-corpus}
\resizebox{\columnwidth}{!}{%
  \begin{tabular}{lrc}
    \toprule
    Corpus & Data Size & Time Taken to Pretrain {\darkbert}\\
    \midrule
    Raw Text & 5.83 GB & 367.4 hours (15.31 days) \\
    Preprocessed Text & 5.20 GB & 361.6 hours (15.07 days) \\
  \bottomrule
\end{tabular}%
}
\end{table}

\begin{figure}
    \centering
    \includegraphics[width=\columnwidth]{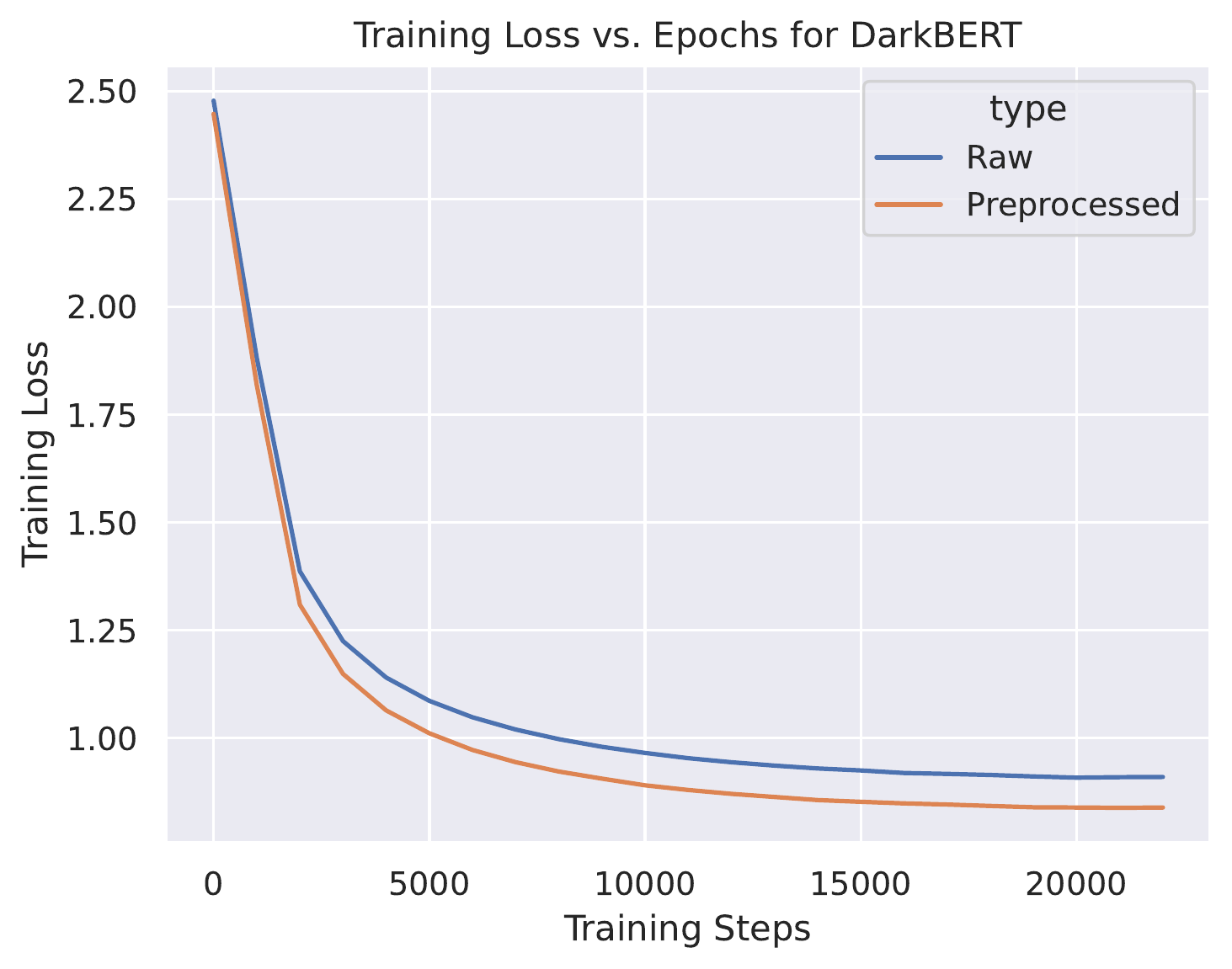}
    \caption{Training steps vs. training loss graph for raw and preprocessed versions of {\darkbert}}
    \label{fig:train-loss}
\end{figure}

\subsection{{\darkbert} Pretraining}
\label{sec:methodology-pretraining}
In order to observe the impact of text preprocessing on DarkBERT's performance, we build two versions of {\darkbert}: one with raw text data (whitespace removal applied) and the other with preprocessed text following Section~\ref{sec:methodology-dataprocess}. The size of each pretraining corpus is shown in Table~\ref{tab:darkbert-pretrain-corpus}, and the training losses for the two models are shown in Figure~\ref{fig:train-loss}.

We leverage an existing model architecture instead of starting from scratch for pretraining. This is done to reduce computational load and retain the general English representation learned by the existing model. We choose RoBERTa~\cite{roberta-paper} as our base initialization model as it opts out of the Next Sentence Prediction (NSP) task during pretraining, which may serve as a benefit to training a domain-specific corpus like the Dark Web as sentence-like structures are not as prevalent compared to the Surface Web.

The Dark Web pretraining text corpus is fed to the \texttt{roberta-base} model in the Hugging Face~\footnote{\url{https://huggingface.co/}} library as an initial base model. For compatibility between {\darkbert} and RoBERTa, we use the same BPE (byte-pair encoding) tokenization vocabulary used in the original RoBERTa model, with each page in the pretraining corpus separated using RoBERTa's separator token \texttt{</s>}. The two versions of {\darkbert} only differ in the corpus used for pretraining (\textit{raw} vs. \textit{preprocessed}); all other factors such as training hyperparameters are equally set. The models are pretrained using a script written in PyTorch~\cite{NEURIPS2019_pytorch}. Additional details on pretraining including hyperparameters and training equipment are listed in Table~\ref{tab:darkbert-hyperparameters} and Section~\ref{sec:app-darkbertpt} of the Appendix.
\section{Evaluation: Dark Web Activity Classification}
\label{sec:evaluation}

In this section, we describe the methods of evaluation and the datasets used to evaluate {\darkbert} and other language models. Since page classification has often been performed in past works~\cite{al-nabki-etal-2017-classifying, choshen-etal-2019-language, ranaldi-darkside}, we also choose Dark Web activity classification as the main Dark Web domain benchmark experiment for evaluation. We additionally conduct experiments on multiple use case scenarios, which is described in detail in Section~\ref{sec:usecase}.

\begin{table}[t]
    \caption{Dataset statistics used for Dark Web activity categorization.}
    \label{tab:dataset-classification}
    \centering
    \resizebox{\columnwidth}{!}{%
    \begin{tabular}{@{}lr|lr@{}}
        \toprule
        \multicolumn{2}{c}{\textbf{DUTA (DUTA-10K)}} & \multicolumn{2}{c}{\textbf{CoDA}}
        \\
        \textbf{Category} & \textbf{Page count} & \textbf{Category} & \textbf{Page count} \\ 
        \midrule
        Hosting \& Software & 1949 & Others & 2131\\
        Cryptocurrency & 798 & Pornography & 1171 \\
        Down & 714 & Drugs & 967 \\
        Locked & 682 & Financial & 956 \\
        Personal & 419 & Gambling & 756 \\
        Counterfeit Credit Cards & 392 & Crypto & 745 \\
        Social Network & 293 & Hacking & 630 \\
        Drugs & 290 & Arms & 597 \\
        Services & 284 & Violence & 482 \\
        Pornography & 226 & Electronics & 420 \\
        Marketplace & 189 & & \\
        Hacking & 182 & & \\
        Forum & 128 & & \\
        \midrule
        Total & 6524 & Total & 8855 \\
        \bottomrule
    \end{tabular}%
    }
\end{table}
\begin{table*}[t]
    \centering
    \caption{Dark Web activity classification evaluation results. Boldface indicates best performance.}
    \label{tab:classification-eval}
    \ra{1.2}
    \resizebox{\textwidth}{!}{%
    	\begin{tabular}{clccc|clccc}
    	\toprule
    	 \textbf{Dataset} & \textbf{Model} &\textbf{Precision} & \textbf{Recall} & \textbf{F1 score} & \textbf{Dataset} & \textbf{Model} &\textbf{Precision} & \textbf{Recall} & \textbf{F1 score} \\ 
    	\midrule    
    	\multirowcell{5}{DUTA\textsubscript{\textit{cased}}} 
    	& BERT\textsubscript{\textit{cased}}   & 77.31 & 76.91 & 77.09 & 
    	\multirowcell{5}{CoDA\textsubscript{\textit{cased}}}
    	& BERT\textsubscript{\textit{cased}} & 92.12 & 92.16 & 92.13
    	\\ 
        & BERT\textsubscript{\textit{uncased}} & 78.21 & 78.20 & 78.19
        & & BERT\textsubscript{\textit{uncased}} & 92.83 & 92.67 & 92.75\\
        & RoBERTa                              & 78.54 & 78.79 & 78.63
        & & RoBERTa & 93.36 & 93.27 & 93.31\\
        & {\darkbert}\textsubscript{\textit{raw}}                          
        & \textbf{80.11} & 79.94 & \textbf{80.01}
        & & {\darkbert}\textsubscript{\textit{raw}} & 94.15 & \textbf{94.35} & 94.25\\
        & {\darkbert}\textsubscript{\textit{preproc}} & 79.90 & \textbf{80.08} & 79.98
        & & {\darkbert}\textsubscript{\textit{preproc}} & \textbf{94.26} & 94.33 & \textbf{94.29}\\
        \midrule
        \multirowcell{5}{DUTA\textsubscript{\textit{uncased}}} 
        & BERT\textsubscript{\textit{cased}}   & 78.11 & 77.97 & 77.99 &
        \multirowcell{5}{CoDA\textsubscript{\textit{uncased}}}
        & BERT\textsubscript{\textit{cased}} & 92.86 & 92.85 & 92.85 \\
        & BERT\textsubscript{\textit{uncased}} & 78.21 & 78.20 & 78.19 
        & & BERT\textsubscript{\textit{uncased}} & 92.83 & 92.67 & 92.75\\
        & RoBERTa                              & 78.42 & 78.36 & 78.37 
        & & RoBERTa & 93.30 & 93.40 & 93.34 \\
        & {\darkbert}\textsubscript{\textit{raw}}                          
        & 79.47 & 79.49 & 79.47
        & & {\darkbert}\textsubscript{\textit{raw}} & \textbf{94.46} & 94.45 & \textbf{94.46} \\
        & {\darkbert}\textsubscript{\textit{preproc}} & \textbf{79.65} & \textbf{79.77} & \textbf{79.69}
        & & {\darkbert}\textsubscript{\textit{preproc}} & 94.31 & \textbf{94.53} & 94.42\\
        \bottomrule
        \end{tabular}
    }
\end{table*}

\subsection{Datasets}
The distribution of various activities has been studied at large, resulting in publicly available Dark Web text datasets known as DUTA~\cite{al-nabki-etal-2017-classifying, ALNABKI2019212} and CoDA~\cite{coda_naacl2022}. We use english texts in the latest version of DUTA (also known as DUTA-10K) and CoDA in our experiments. Since DUTA and CoDA use different categorization methods, we train separate classifiers for each dataset. Since DUTA contains certain categories that are very small in size (for example, there are only 3 pages under the \textit{Human Trafficking} category), we remove categories that have a low page count (under 1\% of total page count). We also remove the \textit{Empty} category in DUTA as these pages are mostly empty, which is not ideal for text classification. No modifications are made to the CoDA dataset. Finally, we preprocess texts in both DUTA and CoDA following Section~\ref{sec:methodology-dataprocess}. Per-category statistics for the two datasets used for our activity classification experiment are shown in Table~\ref{tab:dataset-classification}. 



\subsection{Experimentation}
The classification experiment is conducted on the two versions of {\darkbert} and two widely used language models: BERT~\cite{devlin-etal-2019-bert} and RoBERTa~\cite{roberta-paper}. Although RoBERTa (and the two variants of {\darkbert} which use RoBERTa as their base model) is a cased language model which distinguishes between capitalized words and uncapitalized words, BERT comes in two versions: a cased model and an uncased model. To observe if letter case has any effect on classification performance, we build a separate, \textit{uncased} version of DUTA and CoDA in which every character is converted to lowercase. In summary, we evaluate the Dark Web activity classification task using DUTA and CoDA --- each with two variants: cased and uncased corpus --- on two versions of {\darkbert} (\textit{raw} and \textit{preprocessed}), two versions of BERT (\textit{cased} and \textit{uncased}), and RoBERTa.

\subsection{Results and Discussion}
The result of Dark Web activity classification is shown in Table~\ref{tab:classification-eval}. We observe that DarkBERT outperforms other language models for both datasets and their variants. However, it is also worth noting that both BERT and RoBERTa exhibit relatively similar performances to DarkBERT. This is in line with previous classification experiments with CoDA, which have shown that BERT is able to adapt relatively well to other domains~\cite{coda_naacl2022}. RoBERTa also performs slightly better compared to BERT, which reflects the advantages in performance that RoBERTa has over BERT as mentioned in the original paper~\cite{roberta-paper}. 

We also observe that all language models perform significantly better for the CoDA dataset compared to DUTA. Upon closer inspection on the DUTA dataset, we find that some of the included categories in DUTA may not be suitable for classification tasks. For example, many of the pages in the \textit{Hosting \& Software} category contain duplicate texts, which may overfit the model during fine-tuning (DUTA in general has duplicate texts as mentioned by~\citet{ALNABKI2019212}). In addition, some of the pages seem to be ambiguous in terms of classification the DUTA dataset; for eaxmple, we observe pages classified as \textit{Hosting \& Software} that do not contain any activities related to hosting or software related terms. 

We take a deeper look at the activity classification results on the CoDA (cased) dataset by constructing confusion matrices (Figure~\ref{fig:cm} of the Appendix) to check for misclassifications. We find that in general, the two versions of  {\darkbert} show the best classification performance for most categories. The highest number of correct classifications for every category occurs in either one of the DarkBERT models. However, some categories such as \textit{Drugs}, \textit{Electronics}, and \textit{Gambling} show very similar performances across all four models. This is likely due to the high similarity of pages in such categories, making classification easier even with the differences in the language used in the Dark Web. Finally, we inspect the language models using their predictions through error analysis, which is described in Section~\ref{app:dw-cls} of the Appendix.
\section{Use Cases in the Cybersecurity Domain}
\label{sec:usecase}
In this section, we introduce three Dark Web domain use cases for {\darkbert} and demonstrate its effectiveness over existing language models in cybersecurity / CTI applications. We list details on the experimental setup for each use case in Sections~\ref{app:usecase-leaksite} and \ref{app:usecase-noteworthy} of the Appendix.

\subsection{Ransomware Leak Site Detection}
\label{sec:usecase-leaksite}
			

\begin{figure}[t]
    \begin{subfigure}[b]{\columnwidth}
         \centering
         \includegraphics[width=\textwidth]{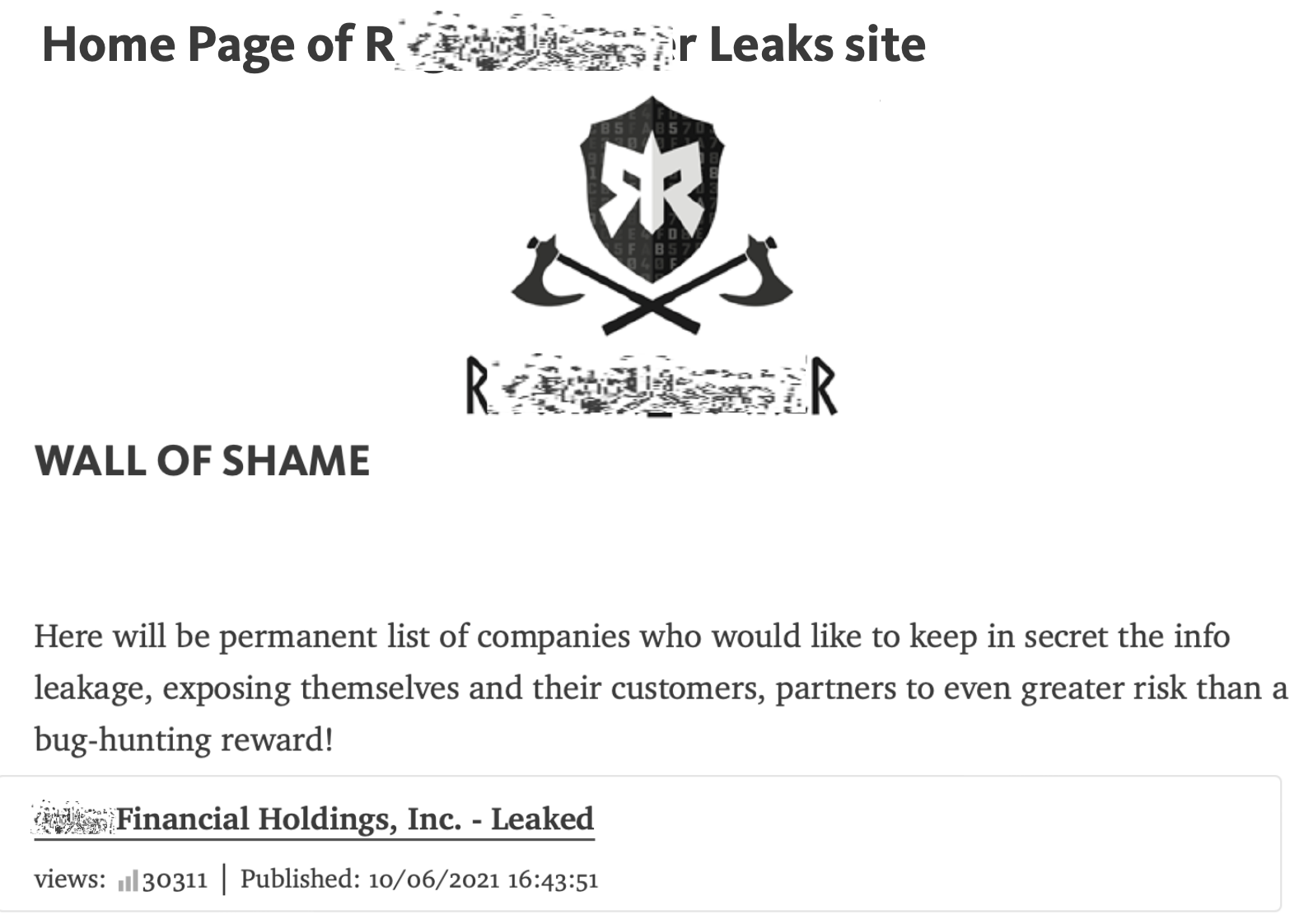}
         \caption{A ransomware leak site sample}
         \label{fig:leak-test-sample}
    \end{subfigure}
    \vfill
    \begin{subfigure}[b]{\columnwidth}
         \centering
         \includegraphics[width=\textwidth]{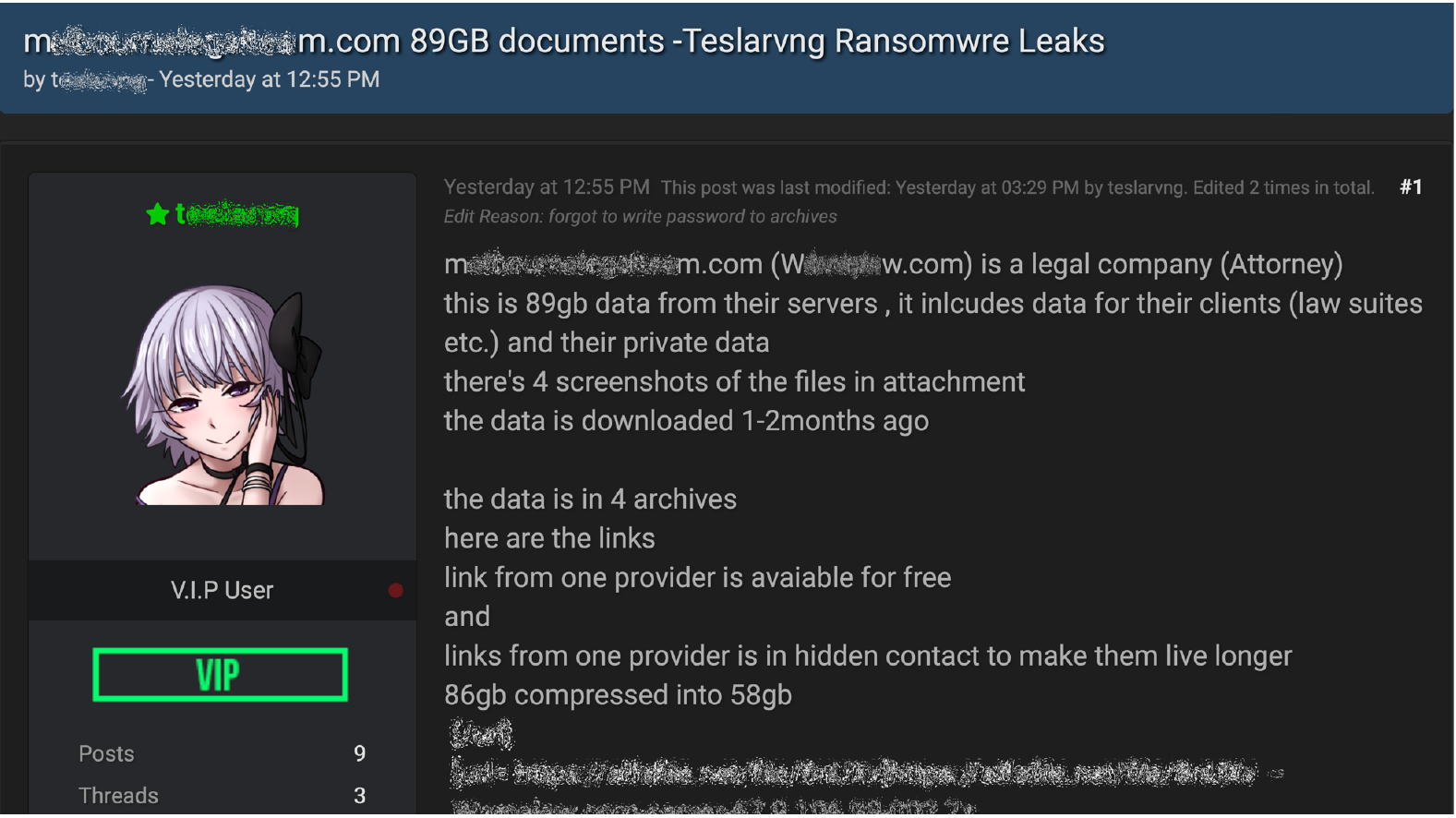}
         \caption{A noteworthy thread sample}
         \label{fig:noteworthy-test-sample}
    \end{subfigure}
    \caption{A ransomware leak site and noteworthy thread samples that {\darkbert} correctly classified but are misclassified by other language models.}
\end{figure}

One type of cybercrime that occurs on the Dark Web is the selling or publishing of private, confidential data of organizations leaked by ransomware groups. 
This can occur in the form of leak sites that expose victims and threaten to release sensitive data (such as financial information, private assets, and personal identification) of uncooperative victims~\cite{YUSTE2021102388}. It would thus be beneficial for security researchers to automatically identify such websites.
We formulate the task of leak site detection as a binary classification problem of predicting whether a given page is a leak site or not. We compare the effectiveness of this task using the pretrained language models used for evaluation in Section~\ref{sec:evaluation} (BERT, RoBERTa, and {\darkbert}).

\noindent \textbf{Datasets}: We monitor leak sites of 54 popular ransomware groups for two years (from May 2020 to April 2022), and periodically download HTML files from these sites especially when new victims are revealed~\footnote{URLs of such leak sites can be found in cybersecurity news, social media, open-source repositories, and so on. We used URLs taken from \url{https://github.com/fastfire/deepdarkCTI/blob/main/ransomware_gang.md.}}. Leak sites typically contain the victim organization name, descriptions of leaked data, and threat statements with sample data (refer to Figure~\ref{fig:leak-test-sample} for an example leak site page). 

We collect pages by randomly choosing a maximum of three pages with different page titles from each of the 54 leak sites, and label them as positive examples. To create negative data, rather than collecting random pages in the Dark Web, we consider pages with content similar to that of leak sites to make the task more challenging. To select such pages, we utilize the activity category classifier from Section~\ref{sec:app-datafiltering} used for balancing the pretraining corpus. The intuition behind using the activity classifier to select negative data is that the text content of certain categories like \textit{Hacking} are more similar to that of leak sites than other less relevant categories such as \textit{Pornography} and \textit{Gambling}. Our pilot study suggests leak sites are mostly classified by the activity classifier as \textit{Hacking}, followed by \textit{Cryptocurrency}, \textit{Financial}, and \textit{Others}. Thus, we only collect Dark Web pages that are classified into one of these four categories and treat them as negative examples. Our training text data consists of 105 positive and 679 negative examples (pages). Training is done using 5-fold cross validation.

\begin{table}[t]
    \centering
    \caption{Ransomware leak site detection performance. Boldface indicates the best performance.}
    \ra{1.2}
    \resizebox{\columnwidth}{!}{%
    	\begin{tabular}{clccc}
    	\toprule
    	 \textbf{Input} & \textbf{Model} &\textbf{Precision} & \textbf{Recall} & \textbf{F1 score} \\ 
    	\midrule    
    	\multirowcell{4}{Raw} 
    	& BERT\textsubscript{\textit{cased}}      & 75.83 & 69.52 & 71.01 \\
        & BERT\textsubscript{\textit{uncased}}    & 77.18 & 73.90 & 72.77 \\
        & RoBERTa                                 & 39.83 & 36.00 & 36.27 \\
        & {\darkbert}\textsubscript{\textit{raw}} & \textbf{78.81} & \textbf{83.62} & \textbf{79.98} \\
        \midrule
        \multirowcell{4}{Preprocessed} 
        & BERT\textsubscript{\textit{cased}}          &	76.81 & 68.19 & 70.13 \\
        & BERT\textsubscript{\textit{uncased}}        & 71.97 & 71.62 & 70.77 \\
        & RoBERTa                                     & 48.36 & 45.14 & 44.31 \\
        & {\darkbert}\textsubscript{\textit{preproc}} & \textbf{85.16} & \textbf{84.57} & \textbf{84.11} \\
        \bottomrule
        \end{tabular}
    }
    \label{tab:leaksite_result}
\end{table}

\noindent \textbf{Results and Discussion}:
As shown in Table~\ref{tab:leaksite_result}, {\darkbert} outperforms other language models, demonstrating the advantages of DarkBERT in understanding the language of underground hacking forums on the Dark Web. Figure~\ref{fig:leak-test-sample} shows a leak site sample correctly classified by {\darkbert} but misclassified by other models. We also observe that while {\darkbert} uses RoBERTa as a base model, RoBERTa itself shows a sharp drop in performance compared to the other models.

In addition, {\darkbert} with preprocessed input performs better than the one with raw input, which highlights the importance of the text preprocessing step in terms of reducing superfluous information. As lengthy words or cryptocurrency addresses have been replaced with mask identifier tokens in the preprocessed input, such words present in the raw input may cause the tokenizer to produce uninformative tokens and affect task performance.

\subsection{Noteworthy Thread Detection}
\label{sec:usecase-noteworthy}


		

\begin{table}[t]
    \centering
    \caption{Noteworthy thread detection performance. Boldface indicates best performance.}
    \ra{1.2}
    \resizebox{\columnwidth}{!}{%
    	\begin{tabular}{clccc}
    	\toprule
    	 \textbf{Input} & \textbf{Model} &\textbf{Precision} & \textbf{Recall} & \textbf{F1 score} \\ 
    	\midrule    
    	\multirowcell{4}{Raw}  
    	& BERT\textsubscript{\textit{cased}}      & 55.09 & 19.91 & 26.90 \\
        & BERT\textsubscript{\textit{uncased}}    & 52.34 & 23.49 & 28.51 \\
        & RoBERTa                                 & 28.97 & 17.89 & 21.38 \\
        & {\darkbert}\textsubscript{\textit{raw}} & \textbf{75.93} & \textbf{43.08} & \textbf{52.85} \\
        \midrule
        \multirowcell{4}{Preprocessed}  
    	& BERT\textsubscript{\textit{cased}}          & 61.43 & 20.48 & 28.81 \\
        & BERT\textsubscript{\textit{uncased}}        & 45.46 & 21.52 & 26.16 \\
        & RoBERTa                                     & 29.04 & 15.27 & 18.71 \\
        & {\darkbert}\textsubscript{\textit{preproc}} & \textbf{72.44} & \textbf{45.13} & \textbf{54.17} \\
        \bottomrule
        \end{tabular}
    }
    \label{tab:noteworthy_result}
\end{table}

Dark Web forums are often used for exchanging illicit information, and security experts monitor for \textit{noteworthy} threads to gain up-to-date information for timely mitigation. Since many new forum posts emerge daily, it takes massive human resources to manually review each thread. Therefore, automating the detection of potentially malicious threads can significantly reduce the workload of security experts. Identifying noteworthy threads, however, requires a basic understanding of Dark Web-specific language. Similar to the aforementioned leak site detection, we can formulate this task as a binary classification problem to predict whether a given forum thread is noteworthy. We compare the performance of noteworthy thread detection for {\darkbert} and the baseline models: BERT and RoBERTa. 

\noindent\textbf{Datasets}:
Identifying a thread as noteworthy is a highly subjective task. While there can be many different definitions for noteworthiness, we focus on activities in hacking forums that can potentially cause damage to a wide range of victims. To incorporate perspectives from the cybersecurity industry and ensure the quality of the dataset, we recruit two researchers from a cyber threat intelligence company specializing in the analysis of hacking forums on the Dark Web to discuss types of noteworthy threads, and set annotation guidelines accordingly. We consider a thread of hacking forums to be \textit{noteworthy} if it describes one of the following activities:

\begin{enumerate}[topsep=0pt,itemsep=-1ex,partopsep=1ex,parsep=1ex]
    \item Sharing of confidential company assets such as admin access, employee or customer information, transactions, blueprints, source codes, and other confidential documents.
    \item Sharing of sensitive or private information of individuals such as credit information, medical records, political engagement, passports, identifications, and citizenship.
    \item Distribution of critical malware or vulnerabilities targeting popular software or organizations.
\end{enumerate}
In particular, we place emphasis on activities targeting large private companies, public institutions, and industries. We choose RaidForums, one of the largest hacking forums, as our data source (together with its mirror and follow-up sites\footnote{\url{http://raidforums.com}, \url{http://rfmirror.com}, \url{http://breached.co}}). We collect 1,873 forum threads posted from July 2021 to March 2022 and work with the recruited annotators to select noteworthy threads. They first annotate the same 150 threads and achieve an inter-annotator agreement of 0.704 as measured by Cohen's Kappa, which indicates \textit{substantial} agreement. All disagreements in the annotated dataset are then discussed and resolved by both annotators. The final dataset contains 249 positive (noteworthy) and 1,624 negative threads. We use the title and body text of each thread from the HTML source as input to the classifier and exclude any thread replies to simulate the practical scenario in which we categorize the noteworthiness of threads as soon as they are posted, and training is done using 5-fold cross validation.

\noindent\textbf{Results and Discussion}:
As seen in Table~\ref{tab:noteworthy_result}, {\darkbert} outperforms other language models in terms of precision, recall, and F1 score for both inputs. Similar to ransomware leak site detection, we see a noticeable performance drop for RoBERTa compared to the other models.
Figure~\ref{fig:noteworthy-test-sample} shows a noteworthy thread sample that is correctly classified by {\darkbert} but misclassified by other models.
Due to the difficulty of the task itself, the overall performance of {\darkbert} for real-world noteworthy thread detection is not as good compared to those of the previous evaluations and tasks. Nevertheless, the performance of {\darkbert} over other language models shown here is significant and displays its potential in Dark Web domain tasks. By adding more training samples and incorporating additional features like author information, we believe that detection performance can be further improved.

It should also be noted that the performances for both raw and preprocessed inputs are similar for {\darkbert}. Unlike data used for ransomware leak site detection, thread content is generally shorter than general webpage content, and sensitive information such as URLs and email addresses often influences the noteworthiness of threads (e.g., whether a victim is a leading global company or not). Since such information is masked for preprocessed inputs, contents of noteworthy threads and non-noteworthy threads may look similar from the viewpoint of the language models, which in turn deteriorates the performance of this task.


\subsection{Threat Keyword Inference}
In this section, we describe how we utilize the fill-mask function to derive a set of keywords that are semantically related to threats and drug sales in the Dark Web. Fill-mask is one of the main functionalities of BERT-family language models, which finds the most appropriate word that fits in the masked position of a sentence (masked language modeling). It is useful for capturing which keywords are used to indicate threats in the wild. In order to show that {\darkbert} is robust in handling this task, we compare {\darkbert} and BERT\textsubscript{Reddit}, a BERT variant fine-tuned on a subreddit corpus whose topic is drugs~\cite{zhu2021selfsupervised}. 

\begin{figure}[tb]
    \centering
    \includegraphics[width=\linewidth]{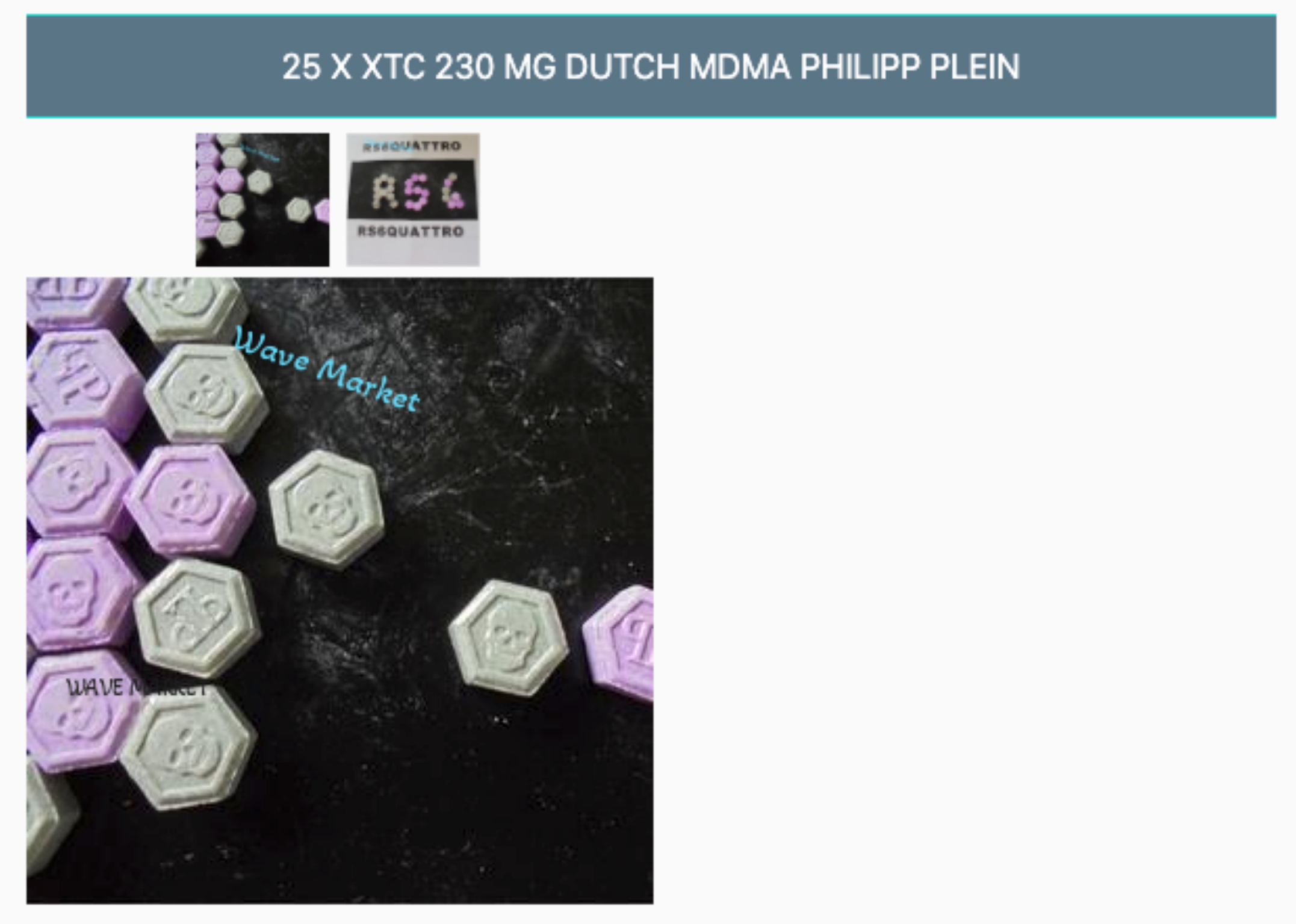}
    \caption{An MDMA sales page excerpted from the Dark Web}
    \label{fig:mdma_pp}
\end{figure}

Figure~\ref{fig:mdma_pp} shows a sample drug sales page from the Dark Web in which a user advertises a Dutch MDMA pill with a \textit{Philipp Plein} logo\footnote{While \textit{Philipp Plein} normally refers to a German fashion brand, in this case, it indicates an MDMA pill on which the brand logo is imprinted. Well known car brands such as \textit{Tesla}, \textit{Rolls Royce}, and \textit{Toyota} are also used in a similar manner.}. We then mask MDMA in the title phrase: \texttt{25 X XTC 230 MG DUTCH MDMA PHILIPP PLEIN}, and let {\darkbert} and BERT\textsubscript{Reddit} suggest the most semantically related words. In Table~\ref{tab:mdma_synonym}, we list the suggested candidate words by the two language models, respectively. The result shows that {\darkbert} suggests drug-related words (i.e., \textit{Oxy} and \textit{Champagne}) and a word closely related to drugs (i.e., \textit{pills}). On the other hand, BERT\textsubscript{Reddit} mainly suggests professions such as \textit{singer}, \textit{sculptor}, and \textit{driver}, which are not relevant to drugs. This comes from the fact that the preceding word, \textit{Dutch}, is usually followed by a vocational word in the Surface Web. We evaluate how each language model produces keyword sets semantically related to drugs in a quantitative fashion.

\begin{table}[tb]
    \centering
    \caption{Fill-mask task results. {\darkbert} suggests specific words related to drugs while BERT suggests general words.}
    \begin{tabularx}{.9\columnwidth}{lX}
        \toprule
        \small \textbf{Language Model} & \small \textbf{Semantically Related Words} \\
        \midrule
        \multirowcell{3}{{\small \darkbert}} & {\small pills, import, md, dot, translation, speed, up, oxy, script, champagne} \\
        \midrule
        \multirowcell{4}{\small BERT\textsubscript{Reddit}} & {\small \#\#man, champion, singer, rider, driver, sculptor, producer, manufacturer, \#\#er, citizen} \\
        \bottomrule
    \end{tabularx}
    \label{tab:mdma_synonym}
\end{table}

\noindent\textbf{Datasets}: To evaluate the language models, we use the sample dataset provided by~\citet{zhu2021selfsupervised}. This dataset is composed of ground truth data (i.e., drug names and their euphemisms) and sentences containing the drug names.~\footnote{The ground truth data are from the DEA Intelligence Report: \url{https://www.dea.gov/sites/default/files/2018-07/DIR-022-18.pdf}}.

\noindent\textbf{Experimental Setting}: We compare three language models: {\darkbert}\textsubscript{CoDA}, BERT\textsubscript{CoDA}, and BERT\textsubscript{Reddit}. The first two language models are fine-tuned on a subset of CoDA documents classified as drugs, whose base model is {\darkbert} and BERT, respectively. BERT\textsubscript{Reddit} is a BERT variant fine-tuned on a subreddit corpus whose topic is drugs. To compare them quantitatively, we also use precision at $k$ ($P@k$) following~\citet{zhu2021selfsupervised}. Here, precision at $k$ is the proportion of inferred keywords that are semantically related to a given drug name in the top-$k$ set that are synonymous.

\noindent\textbf{Results and Discussion}: The measured $P@k$ values are presented in Table~\ref{tab:usecase_3_eval}. {\darkbert}\textsubscript{CoDA} outperforms BERT\textsubscript{Reddit} for $k$ ranging from 10 to 20, but is overtaken for higher values of $k$. Although {\darkbert}\textsubscript{CoDA} shows better performance when $k$ is small, the ground truth dataset contains euphemisms mainly derived from the Surface Web, and the words that {\darkbert}\textsubscript{CoDA} infers as semantically related words are not contained in the dataset. For instance, \textit{Tesla} and \textit{Champagne} are drug names frequently seen in the Dark Web, but are not recognized as such in~\citet{zhu2021selfsupervised}. On the other hand, \textit{crystal} and \textit{ice} are detected by both {\darkbert}\textsubscript{CoDA} and BERT\textsubscript{Reddit} because they are used in both the Surface Web and the Dark Web.


\begin{table}[t]
    \centering
    \caption{Quantitative performance metric of threat keyword inference. Precision at $k$ ($P@k$) is measured with varying $k$ in increments of 10.}
    \resizebox{\columnwidth}{!}{%
        \begin{tabular}{cccccc} \toprule
                                       & Top-10 & Top-20 & Top-30 & Top-40 & Top-50 \\ \midrule
             {\darkbert}\textsubscript{CoDA} & \textbf{0.60} & \textbf{0.60} & 0.50 & 0.42 & 0.42 \\
             BERT\textsubscript{CoDA}        & 0.40 & 0.40 & 0.50 & 0.50 & 0.40 \\
             BERT\textsubscript{Reddit}      & 0.40 & 0.45 & \textbf{0.60} & \textbf{0.57} & \textbf{0.52} \\ \bottomrule
        \end{tabular}
    }
    \label{tab:usecase_3_eval}
\end{table}

\section{Conclusion}
In this study, we propose {\darkbert}, a Dark Web domain-specific language model based on the RoBERTa architecture.
To allow {\darkbert} to adapt well to the language used in the Dark Web, we pretrain the model on a large-scale Dark Web corpus collected by crawling the Tor network. We also polish the pretraining corpus through data filtering and deduplication, along with data preprocessing to address the potential ethical concerns in Dark Web texts related to sensitive information. 
We show that {\darkbert} outperforms existing language models with evaluations on Dark Web domain tasks, as well as introduce new datasets that can be used for such tasks. {\darkbert} shows promise in its applicability on future research in the Dark Web domain and in the cyber threat industry. In the future, we also plan to improve the performance of Dark Web domain specific pretrained language models using more recent architectures and crawl additional data to allow the construction of a multilingual language model.
\clearpage
\section*{Ethical Considerations}
\label{sec:ethics}

\subsection*{Crawling the Dark Web}
While crawling the Dark Web, we take caution not to expose ourselves to content that should not be accessed. For example, illicit pornographic content (such as child pornography) are easily found on the Dark Web. However, our automated web crawler takes the approach of removing any non-text media and only stores raw text data. By doing so, we do not expose ourselves to any sensitive media that is potentially illegal. 

\subsection*{Sensitive Information Masking}
Since the Dark Web harbors many activities considered to be malicious in nature, it is of utmost importance that sensitive data be left out of the text corpus used for pretraining. In particular, it is possible that some contents in the Dark Web may include private information such as e-mails, phone numbers, or IP addresses. To prevent {\darkbert} from learning representations from sensitive texts as mentioned above, we mask our data before feeding it to our language model. While we have used both {\darkbert} pretrained on preprocessed text and raw text for our experiments, we have used both of the models only for evaluation purposes. In addition, we plan to only release the preprocessed version of {\darkbert} in order to avoid any malpractices once the model is made publicly available~\footnote{More information on the sharing of the datasets and the model itself will be released during the conference.}. Through extensive testing on fill-mask and synonym inference tasks, we observe that it is infeasible to infer any characteristics or data that might be considered sensitive or private in nature using the preprocessed version of {\darkbert}.

\subsection*{Annotator Ethics}
For the task of noteworthy thread detection, we recruited two researchers from a cyber threat intelligence company as mentioned in Section~\ref{sec:usecase-noteworthy}, who agreed to assist us in our research methods. For a fair annotation process in the discussion of noteworthy threads, both recruited annotators handled the same set of thread data and were given equal compensations. 

\subsection*{Use of Public Dark Web Datasets}
Both DUTA and CoDA are available upon request by the respective authors, and due to the sensitive nature of the Dark Web domain, these datasets are only to be used for academic research purposes. We adhere to this guideline and only utilize the provided data in the context of research for this work. On the other hand, we do not plan to publicly release the Dark Web text corpus used for pretraining DarkBERT for similar reasons.
\section*{Limitations}
\label{sec:limitations}

\subsection*{Limited Usage for Non-English Tasks}
As mentioned in Section~\ref{sec:methodology}, DarkBERT is pretrained using Dark Web texts in English. This is mainly our design choice as the vast majority (around 90\%) of Dark Web texts is primarily in English \citep{coda_naacl2022}. We believe that with the limited number of collected pages in non-English languages in our pretraining corpus, building a multilingual language model for the Dark Web domain would pose additional challenges, such as downstream task evaluations becoming more difficult to perform as they would require high-quality annotations of task-specific datasets in multiple languages. As such, while our language model is suitable for Dark Web tasks in English, further pretraining with language-specific data may be necessary to use DarkBERT for non-English tasks.

\subsection*{Dependence on Task-Specific Data}
Although DarkBERT is a useful tool that can be directly applied to many existing Dark Web domain-specific tasks, some tasks may require further fine-tuning through task-specific data (as seen in Ransomware Leak Site Detection and Noteworthy Thread Detection use case scenarios in Section~\ref{sec:usecase}). However, there is a shortage of publicly available Dark Web task-specific data. While we provide the datasets used to fine-tune DarkBERT in this paper, additional research on tasks that do not have readily available datasets for use may require further manual annotation or handcrafting of necessary data to leverage DarkBERT to its maximum potential. 

\section*{Acknowledgements}
\label{sec:ack}

This work was supported by Institute of Information \& Communications Technology Planning \& Evaluation (IITP) grant funded by the Korea government (MSIT). (No.2022-0-00740, The Development of Darkweb Hidden Service Identification and Real IP Trace Technology)


\bibliography{anthology,custom,biblio}
\bibliographystyle{acl_natbib}
\balance

\appendix

\clearpage
\section{Appendix}
\label{sec:appendix}

We list some additional details such as example figures from the {\darkbert} evaluation and cybersecurity use case experiments mentioned in Sections~\ref{sec:evaluation} and~\ref{sec:usecase}. Select portions of figures have been blurred out to comply with the ethical guidelines to hide sensitive information.

\section{Data Filtering Details}
\label{sec:app-datafiltering}

\noindent\textbf{Removal of pages with low information density}:
Initially, we decide to leave out pages that have an abnormally high or low character count. This is done to exclude content that is not seemingly useful in the representation of the Dark Web. For example, most of the pages containing an abnormally low character count are error messages such as ``404 not found'' or ``Captcha error'' and log-in messages such as ``Sign In'' or ``Already have an account?''. On the other hand, the pages that contain an abnormally high character count are mostly large lists of keywords or continuous repetitions of certain strings. These texts are not very useful as they contain low information density of Dark Web content, and are therefore removed from the pretraining corpus.

To decide on the minimum and the maximum threshold of character counts to remove from the crawled data, we measure the per-page character count statistics as shown in Table~\ref{tab:darkweb-data-stats}, and use approximately half the character count value from the 25th quartile (500 characters) and double the character count value from the 75th quartile (10,000 characters). This is done so that the majority of the pages are still included in the pretraining corpus while also serving as a generous threshold for pages containing unwanted data as shown above. By filtering out pages below the minimum and above the maximum threshold for their character count, we are left with 5.43 million pages out of the initial 6.1 million.

\begin{table}[b]
\centering
\caption{Dark Web data collection statistics}
\label{tab:darkweb-data-stats}
\resizebox{0.9\columnwidth}{!}{%
  \begin{tabular}{cc}
    \toprule
    Statistics & Value\\
    \midrule
    Total number of collected pages & 6.1 M \\
    Average number of characters per page & 7,980 \\
    Minimum number of characters in a page & 7\\
    Maximum number of characters in a page & 17,786,986\\
    \midrule
    Per-page character count statistics & Character count \\
    \midrule
    $Q_1$ (25th quartile) & 1,318\\
    $Q_2$ (50th quartile) & 2,581\\
    $Q_3$ (75th quartile) & 5,753\\
  \bottomrule
\end{tabular}%
}
\end{table}

\noindent\textbf{Category balancing}:
Previous studies~\cite{al-nabki-etal-2017-classifying, coda_naacl2022} have found through their web crawling that pornographic content is one of the most common activities found in the Dark Web. One of the challenges in pretraining {\darkbert} is to use text data that consists of various content found in the Dark Web while avoiding skewness in which certain activities constitute a significant fraction of the entire dataset. If these activities (that take up a large portion of the corpus) exist, then the learned representation of the language model would be more biased towards such activities through pretraining.

To address the issue of balancing content in the pretraining corpus, we attempt an automated categorization of every page. A general categorization of various activities and the guidelines for each activity were addressed by~\citet{coda_naacl2022}, where each page in the Dark Web was sorted into a total of 10 categories. Following this classification methodology, we train a simple page classification model using BERT~\cite{devlin-etal-2019-bert}. Although the use of vanilla BERT may seem contradicting due to the domain differences between the Surface Web (the domain of origin for the texts that BERT was pretrained with) and the Dark Web, it is not necessary for this classification model to achieve high performance since our goal is to obtain a general grasp of the pretraining corpus category distribution. 

We implement the model by finetuning the \texttt{bert-base-uncased} model from the Hugging Face library~\cite{wolf-etal-2020-transformers} with the CoDA Dark Web text corpus~\cite{coda_naacl2022}. This model is then run through the entire pretraining corpus to output a specific category for each page. We use 9 of the 10 predefined categories from CoDA and exclude the \textit{Others} category, because most of the pages that fit in this category (log-in pages, error pages, etc.) have already been filtered out from the pretraining corpus through character count filtering. In addition, we found that pages are more likely to be misclassified as \textit{Others} category compared to other categories, meaning that the exclusion of \textit{Others} category would yield a more accurate category distribution.

\begin{table*}[t]
\centering
\caption{Dark Web page classification and pretraining data statistics. The statistics marked as \textit{(full)} represent the original data collection, and \textit{(pretraining)} represents the data after deduplication and category balancing are applied.}
\label{tab:darkweb-data-all}
\resizebox{\linewidth}{!}{%
  \begin{tabular}{lrcc|rcrr}
    \toprule
    Category & Page Count (full) & Total Size (full) & Average Size per Page (full) &
               Page Count (pretraining) & Total Size (pretraining) & Deduplication Rate & Total Reduction Rate\\
    \midrule
    Pornography & 2,267,628 & 9.70 GB & 4.28 KB & 224,781 & 971.0 MB & 2.91\% & 89.98\% \\
    Drugs & 503,433 & 1.75 GB & 3.47 KB & 228,965 & 766.7 MB & 23.31\% & 56.19\% \\
    Financial & 637,917 & 2.10 GB & 3.29 KB & 253,171 & 874.1 MB & 12.45\% & 58.38\% \\
    Gambling & 43,041 & 0.15 GB & 3.38 KB & 40,584 & 137.5 MB & 5.37\% & 5.37\% \\
    Cryptocurrency & 412,349 & 1.36 GB & 3.29 KB & 249,811 & 897.6 MB & 10.28\% & 34.00\% \\
    Hacking & 801,330 & 3.51 GB & 4.38 KB & 57,183 & 242.7 MB & 75.73\% & 93.09\% \\
    Arms / Weapons & 46,616 & 0.14 GB & 2.70 KB & 43,250 & 129.9 MB & 6.15\% & 6.15\% \\
    Violence & 323,738 & 1.21 GB & 3.74 KB & 253,566 & 959.8 MB & 4.02\% & 20.68\% \\
    Electronics & 401,196 & 0.89 GB & 2.21 KB & 381,218 & 850.4 MB & 4.17\% & 4.45\% \\
    \midrule
    Total & 5,437,248 & 20.79 GB & - & 1,732,529 & 5.83 GB & 18.69\% & 71.96\% \\
  \bottomrule
\end{tabular}%
}
\end{table*}

The page category statistics resulting from classification is shown in Table~\ref{tab:darkweb-data-all}. We observe from our data that \textit{pornography} accounts for the highest fraction of all categories in the Dark Web, making up 41.7\% of all pages. Meanwhile, categories such as \textit{gambling} and \textit{arms / weapons} make up less than 1\% of all pages each. Even with the use of vanilla BERT and the exclusion of the \textit{Others} category taken into consideration, it is evident that the variation of content in the pretraining corpus is unbalanced. To this end, we take a rather simple approach of random removal of pages from over-represented categories until all categories have similar amounts of content. 


\noindent\textbf{Deduplication}:
A significant portion of the Dark Web is duplicate content. Since pretraining language models requires considerable resource and time, reducing the pretraining corpus size through deduplication is beneficial. This process is handled by minhashing~\cite{BRODER2000630} each page in the corpus and removing duplicate pages until all remaining minhash values are unique.

The pretraining corpus statistics after applying random removal of over-represented pages and deduplication is shown in Table~\ref{tab:darkweb-data-all}. The deduplication rate represents the reduction in data size as a result of deduplication only, while the total reduction rate represents the reduction in data size as a result of both deduplication and random removal for category balancing. Both are based on the ratio between the initial data size (Table~\ref{tab:darkweb-data-all}) and the final data size of each category. We observe that most categories have deduplication rates of less than 10\%. However, categories such as \textit{drugs} and \textit{hacking} exhibit high deduplication rates. In addition, the deduplication rate and the total reduction rate of \textit{gambling} and \textit{arms / weapons} categories are equal, since we did not perform random removal of pages as these categories were already initially small in terms of data size. Finally, the size difference between the smallest category (\textit{arms / weapons}) and the largest category (\textit{pornography}) is 7-fold in the final pretraining corpus, compared to the 70-fold difference in size observed from the initial data. 

\section{Identifier Mask Details}
\label{sec:app-idmask}

\begin{table*}
\centering
\caption{The types of identifier masks and the list of preprocessed texts.}
\label{tab:preprocessing-type}
\resizebox{\linewidth}{!}{%
  \begin{tabular}{llcl}
    \toprule
    Identifier Type & Example Text or Description & Preprocess Action Type & Identifier Mask Token\\
    \midrule
    Email Addresses & example@email.com & Replace with token & \texttt{ID\_EMAIL} \\
    URLs (non-onion domain) & www.example.com & Replace with token & \texttt{ID\_NORMAL\_URL} \\
    & https://www.example.com/home & & \\
    URLs (onion domain) & facebookwkhpilnemxj7asaniu7vnjjbiltxjqhye3mhbshg7kx5tfyd.onion & Replace with token & \texttt{ID\_ONION\_URL}\\
    IP Addresses (IPv4 \& IPv6) & \texttt{192.168.1.1} & Replace with token & \texttt{ID\_IP\_ADDRESS}\\
    & \texttt{fe80::1ff:fe23:4567:890a\%eth2} & & \\
    Cryptocurrency Addresses & BTC, ETH, LTC addresses & Replace with token & \texttt{ID\_BTC\_ADDRESS}\\
    & & & \texttt{ID\_ETH\_ADDRESS}\\
    & & & \texttt{ID\_LTC\_ADDRESS}\\
    Lengthy ``Words'' & Any group of non-whitespace characters that are 38 or more letters long & Replace with token & \texttt{ID\_LONGWORD}\\
    Uncommon Characters & Any characters out of Unicode range from \texttt{U+0000} to \texttt{U+00FF} & Remove from text & -\\
    Whitespaces & Newline characters, tabs, spaces, etc. & Truncate to a single space & -\\
  \bottomrule
\end{tabular}%
}
\end{table*}

Here, we give an extended discussion on each of the identifier masks used for text processing mentioned in Section~\ref{sec:methodology-dataprocess}. The types of identifier masks used for preprocessing the pretraining corpus is illustrated in Table~\ref{tab:preprocessing-type}.

\noindent\textbf{Implementation}: Some identifier types such as URLs and IP addresses always contain distinct patterns. These identifiers are searched and undergo substitution using regular expressions. Other identifier types such as emails and phone numbers are masked using the text preprocessing API provided by textacy~\footnote{\url{https://textacy.readthedocs.io/en/latest/}}.

\noindent\textbf{Email Addresses}: Email addresses are often seen in the Dark Web as a means of communication. Unlike the contacts commonly seen in the Surface Web, many of the email addresses listed in the Dark Web are those that provide end-to-end encryption services such as ProtonMail~\footnote{\url{https://proton.me/mail}} to prioritize privacy. However, some email addresses can include strings that can be traced to a single individual, so all email addresses are masked.

\noindent\textbf{URLs}: There are two identifier types for URLs: onion domain addresses and non-onion domain addresses. While URLs do not necessarily expose personal information themselves, it is possible that links to some URLs may be contain harmful information or data. To eliminate the possibility of such URLs from being learned as a representation of the Dark Web, we mask all URLs.

\noindent\textbf{IP Addresses}: Although the Dark Web is used to hide IP addresses, some pages contain IP addresses in their texts. Many of the pages that contain IP addresses are Tor relay sites, which show information such as Tor exit relay node addresses (the IP addresses listed in the Tor relay sites can also easily be found on the Surface Web at \url{https://metrics.torproject.org/rs.html}). Given the frequent illegal activities occurring in the Dark Web, it is possible that some IP addresses listed in these pages may exist for malicious purposes. For example,~\citet{exit-relays} has shown that some malicious exit relays have been engaging in HTTPS man-in-the-middle attacks. Therefore, we found it necessary to mask all IP addresses (both IPv4 and IPv6 addresses are masked).

\noindent\textbf{Lengthy Words}: While exploring some of the unpreprocessed text in the pretraining corpus, we found that certain pages contain words (string of characters separated by whitespace) that are extremely long in length. On closer inspection, most of these lengthy words are URLs, code snippets, hash values, file names, cryptocurrency addresses, and even binaries. While URLs and cryptocurrency addresses can be removed through the preprocessing mask identifiers, other types such as hashes and file names are not separately processed in advance. Hashes in particular would incur overhead in building meaningful vocabulary through tokenization as they do not have specific lexical patterns. In addition, since we do not want executable content such as binaries or detailed file names to be learned by our language model, we decide to mask all lengthy words. To this end, we define \textit{lengthy words} by studying the word length distribution, as well as manual inspection of example words for some notable word lengths. 


\begin{figure}[t]
    \centering
    \includegraphics[width=\columnwidth]{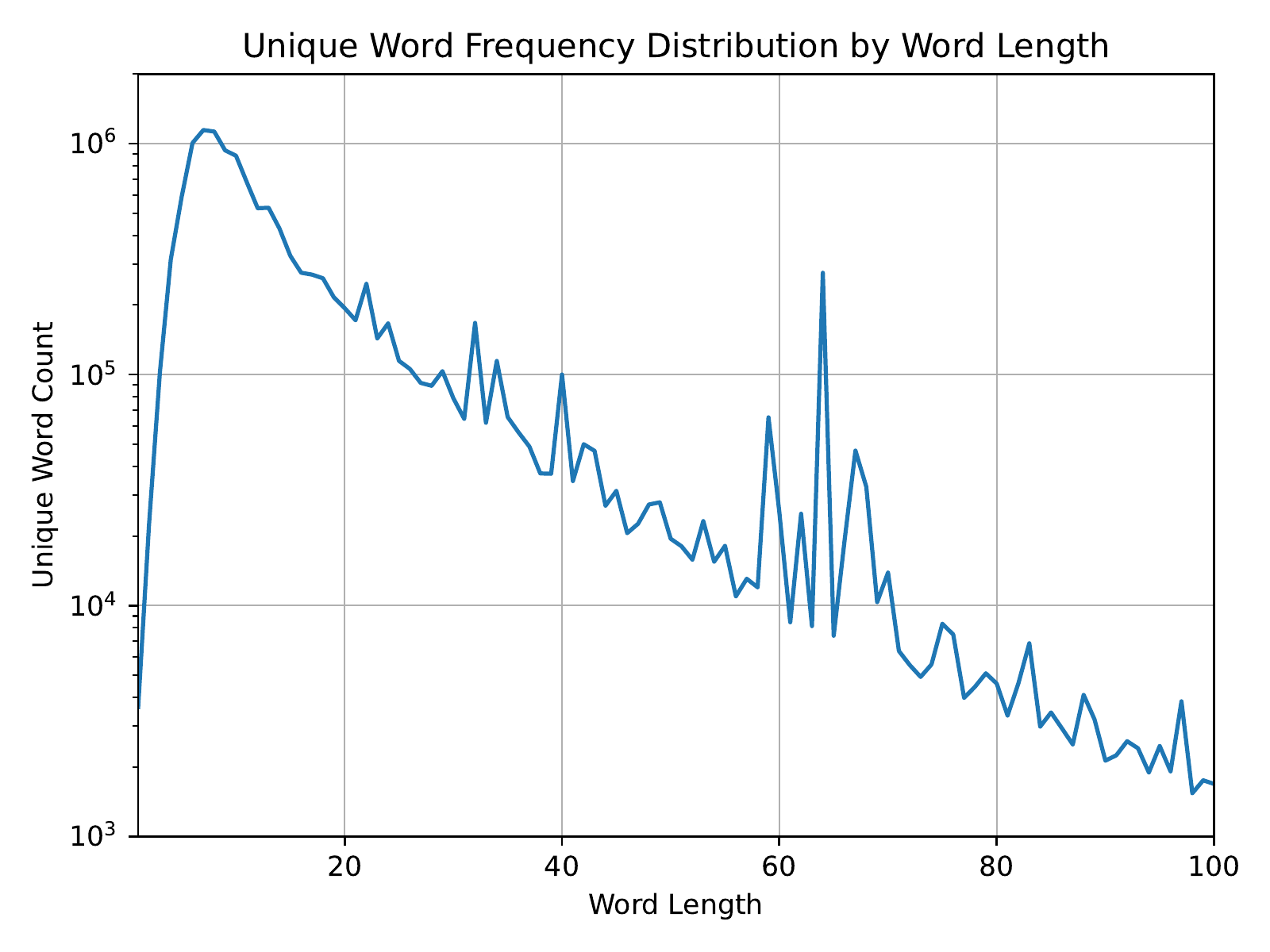}
    \caption{Unique word length distribution for the pretraining corpus before preprocessing is applied. Word lengths greater than 100 are omitted for brevity.}
    \label{fig:uniq_word_dist}
\end{figure}

The unique word length distribution for the pretraining corpus is shown in Figure~\ref{fig:uniq_word_dist}. The word length distribution shows a steep upward trend at shorter word lengths (peaking at length of 7) similar to the English language word distribution, and gradually decreases with longer word lengths. As observed in the figure, some specific word lengths appear in much greater frequencies at higher levels. Upon inspection, we find that this is due to some of the commonly used string formats that happen to have specific lengths. For example, many words of length 59 found in our corpus are content identifier (CIDv1) hashes commonly used in IPFS~\footnote{https://ipfs.io/}, which is a decentralized, hypermedia distribution protocol. Similarly, words of length 64 are mostly SHA-256 hashes.


Our manual inspection of some of the vocabularies present for each word length shows that at around length of 38 to 40, the majority of words take the form of hash-like values and meaningless noisy strings. Therefore, we classify words with lengths of 38 or more characters as \textit{lengthy words}, and mask them from the pretraining corpus. Note that the masking process of \textit{lengthy words} is performed after masking all other identifiers mentioned previously such as email addresses, URLs, and cryptocurrency addresses. Since texts belonging to such identifiers are lengthy (ex. onion V3 addresses are 56 characters long, and Ethereum addresses consist of 40 digit hexadecimal strings), masking these texts with their associated mask identifiers beforehand prevents them from being misclassified as \textit{lengthy words}.

\noindent\textbf{Uncommon Characters}: As mentioned in Section~\ref{sec:methodology-data-collection}, we collect pages that are classified as ``English''. However, some of these collected pages contain multilingual characters that are not standard English. The inclusion of such nonstandard characters results in noisy tokens during the tokenization process and produces unnecessary token vocabularies, so we remove all the characters that are ``uncommon'' in contemporary English. Specifically, we remove all Unicode characters that are not one of the 256 characters in the \texttt{Basic Latin} (ASCII characters) and the \texttt{Latin-1 Supplement} (accented alphabets that are often seen in English) category.

\noindent\textbf{Cryptocurrency Addresses}: Decentralized digital assets like cryptocurrencies are used to make unidentifiable transactions. As many cryptocurrencies are secure by design and provide pseudonymity, the synergy with the anonymous nature of the Dark Web makes them the preferred method of choice for transactions. Studies show that cryptocurrencies have been involved in illegal underground operations~\cite{cybercriminal-minds} in the Dark Web and underground marketplaces in general~\cite{soska-190886}. While cryptocurrencies are known for their pseudonymous properties, many of the transactions are traceable as the entire blockchain is public (for some cryptocurrencies). In particular, we mask Bitcoin, Ethereum, and Litecoin addresses as these three cryptocurrencies are among the most popular in the Dark Web with transparent transaction details (Monero and Dash are also popular in the Dark Web, but they incorporate added layers of anonymity to further conceal their transactions).~\cite{barysevich2018litecoin}.

\begin{table}[t]
    \centering
    \caption{The hyperparameters used for pretraining the two versions of DarkBERT.}
    \label{tab:darkbert-hyperparameters}
    \resizebox{0.5\columnwidth}{!}{%
      \begin{tabular}{lc}
        \toprule
        Hyperparameter & Value\\
        \midrule
        Number of Layers & 12  \\
        Hidden Size & 768 \\
        Feedforward NN & \\ 
        Inner Hidden Size & 3072 \\
        Attention Heads & 12 \\
        Attention Head Size & 64 \\
        Dropout & 0.1 \\
        Attention Dropout & 0.1 \\
        Max Sequence Length & 512 \\
        Warmup Steps & 24000 \\
        Peak Learning Rate & 6e-4 \\
        Batch Size & 8192 \\
        Weight Decay & 0.01 \\
        Max Steps & 20K \\
        Learning Rate Decay & Linear \\
        Adam $\epsilon$ & 1e-6 \\
        Adam $\beta_{1}$ & 0.9 \\
        Adam $\beta_{2}$ & 0.98 \\
        Gradient Clipping & 0.0 \\
      \bottomrule
    \end{tabular}%
    }
\end{table}

\section{DarkBERT Pretraining Details}
\label{sec:app-darkbertpt}


Both versions of {\darkbert} are pretrained on a machine with Intel Xeon Gold 6348 CPU @ 2.60GHz and 4 NVIDIA A100 80GB GPUs. All 4 GPUs were used to run the pretraining process, and each version of {\darkbert} took about 15 days to run (up to 20K training steps --- we stopped the pretraining process at training loss convergence). Both versions of {\darkbert} share relatively similar training losses over the 20K training steps. Since training loss for both versions of {\darkbert} stopped decreasing at around 20K steps, we use the models saved at 20K steps for evaluation. 

\section{Evaluation Details}
\label{sec:app-eval}



\subsection{Dark Web Activity Classification}
\label{app:dw-cls}
We implement a classification pipeline using the language models available in the Hugging Face library (\texttt{bert-base-cased}, \texttt{bert-base-uncased}, and \texttt{roberta-base}) and add a fully-connected classification layer on top of the \texttt{[CLS]} token with PyTorch. Evaluation is performed for each model using $k$-fold cross validation ($k = 10$), which is implemented using scikit-learn's \texttt{StratifiedKFold} module~\cite{scikit-learn}. Each fold is run up to 10 epochs with a learning rate of \texttt{2e-5}.

\noindent\textbf{Error Analysis}: We further scrutinize model performance by looking at specific pages in the CoDA dataset that are correctly classified by {\darkbert} but are misclassified by the other models. We find that most pages that have been misclassified by BERT and RoBERTa but correctly classified by DarkBERT contain many domain-specific jargons or key phrases seen in that particular activity in the Dark Web. For example, one of the pages under the \textit{Financial} category that is misclassified by both BERT and RoBERTa as \textit{Others} contains the name of a credit card seller service (we choose not to reveal the service name for ethical considerations). Another page under the \textit{Pornography} category contains the phrase \textit{red room} which is highly correlated to this category of pages in the Dark Web, but is misclassified by both BERT and RoBERTa as \textit{Others}. Finally, a page under the \textit{Crypto} category contains blockchain and cryptocurrency terms, but is misclassified by BERT and RoBERTa as \textit{Others}. As shown in the above examples, DarkBERT is able to correctly classify pages that contain phrases mostly seen in the Dark Web but are not commonly used in the Surface Web, whereas BERT and RoBERTa tend to misclassify such pages in the \textit{Others} category as these models consider such words and phrases as generic attributes rather than activity-specific terms.

\begin{figure}[H]
    \centering
    \includegraphics[width=.95\linewidth]{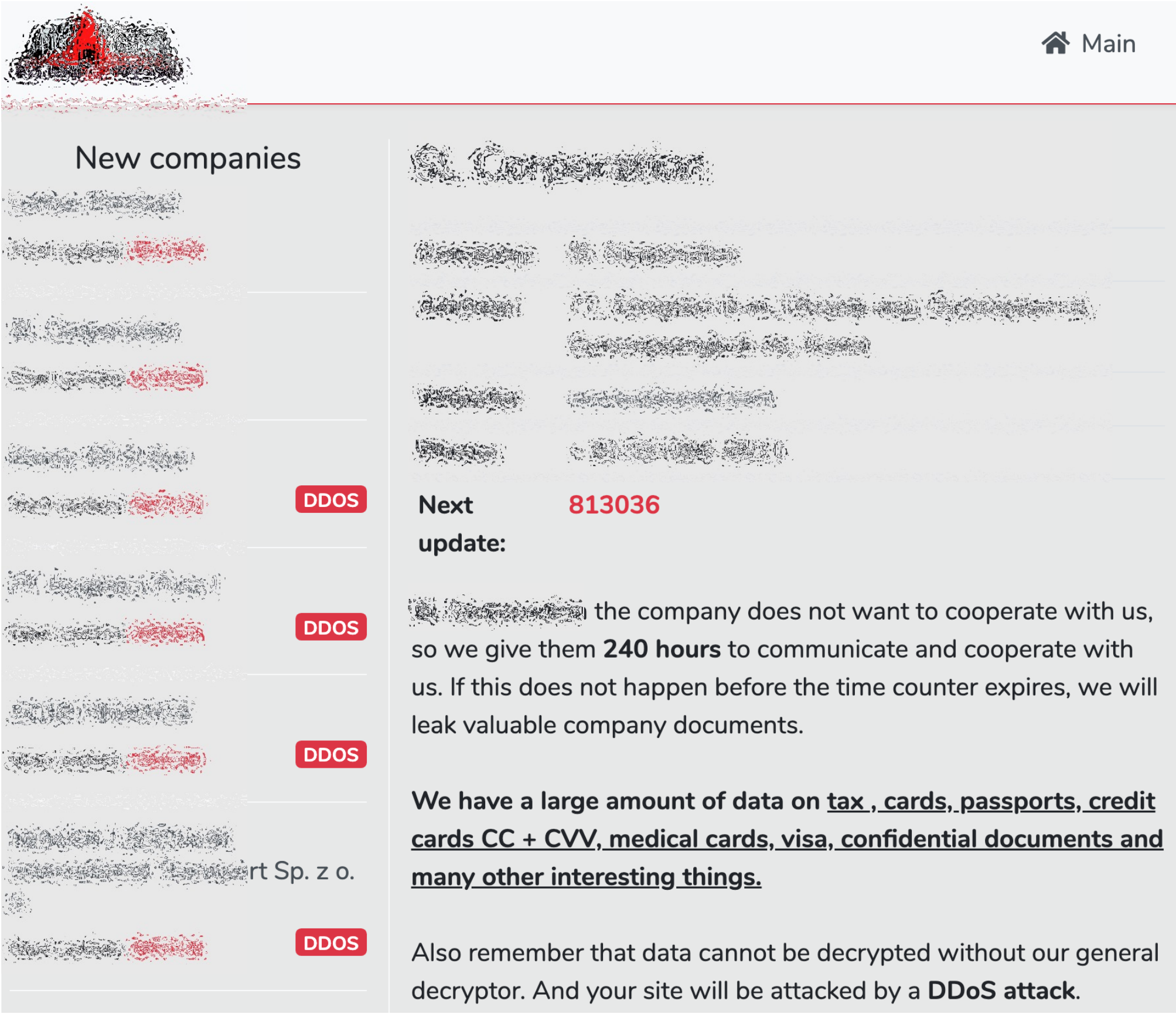}
    \caption{A leak site page sample in the dataset.}
    \label{fig:usecase-leaksite-samples}
\end{figure}

\subsection{Ransomware Leak Site Detection}
\label{app:usecase-leaksite}

We use the same classification pipeline as activity classification in Section \ref{sec:evaluation} with $k$-fold cross-validation ($k = 5$) and connect fully-connected classification layers on top of the \texttt{[CLS]} token. Similarly, the evaluation is performed on both raw and preprocessed inputs. An early stopping strategy using validation loss is utilized to avoid overfitting. Due to the limited size of the dataset, we choose to repeat $k$-fold validation 5 times to mitigate the variations in performance per run and average the results. An example data sample used for this task can be seen in Figure~\ref{fig:usecase-leaksite-samples} and additional details on used hyperparameters can be found in Table~\ref{tab:usecase-hyperparameters}.

\begin{figure}[H]
    \centering
    \includegraphics[width=.95\linewidth]{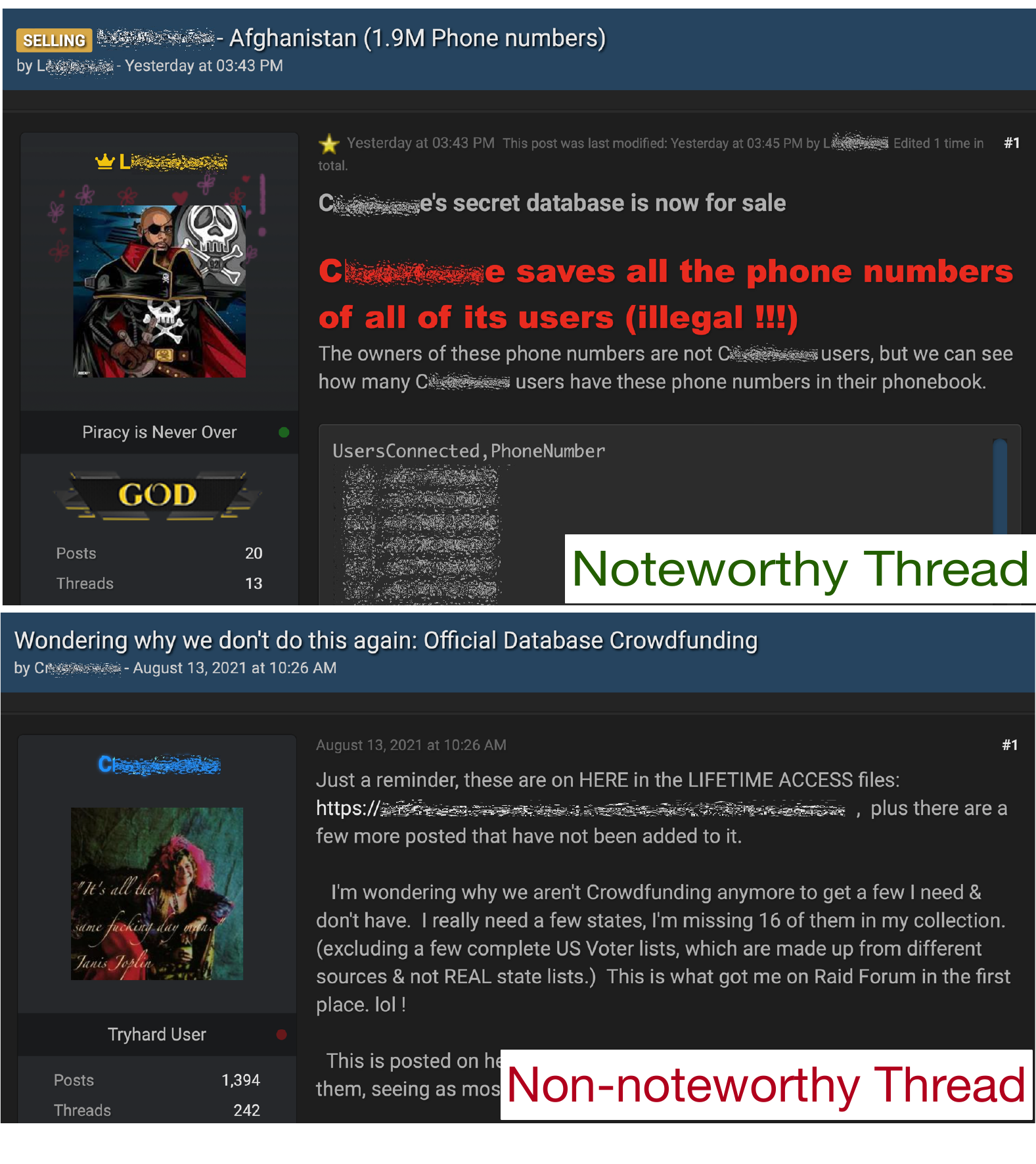}
    \caption{Noteworthy and non-noteworthy thread samples in the dataset.}
    \label{fig:usecase-noteworthy-samples}
\end{figure}

\subsection{Noteworthy Thread Detection}
\label{app:usecase-noteworthy}

Similar to ransomware leak site detection, we adopt $k$-fold cross validation ($k = 5$) for each model and employ early stopping strategy. Due to the limited size of the dataset, we again use repeated $k$-fold validation, where the number of repetitions is set to 5. An example data sample used for this task can be seen in Figure~\ref{fig:usecase-noteworthy-samples} and additional details on used parameters can be found in Table~\ref{tab:usecase-hyperparameters}.

\begin{table}[ht]
\centering
\caption{The hyperparameters used in ransomware leak site detection and noteworthy thread detection.}
\label{tab:usecase-hyperparameters}
\resizebox{0.95\columnwidth}{!}{%
  \begin{tabular}{lcc}
    \toprule
    \multirow{2}{*}{Hyperparameter} & Ransomware leak site & Noteworthy thread \\
    & detection & detection \\
    \midrule
    Epochs & 100 & 100 \\
    Batch Size & 32  & 32\\
    Learning Rate & 1e-4 & 1e-5\\
    Number of Layers & 2 & 2\\
    Hidden Size & 64 & 64 \\
    Dropout & None & 0.5 \\
  \bottomrule
\end{tabular}%
}
\end{table}

\begin{figure*}[ht]
     \centering
     \begin{subfigure}[b]{0.48\textwidth}
         \centering
         \includegraphics[width=\textwidth,trim={0 0 1.5cm 1.5cm},clip]{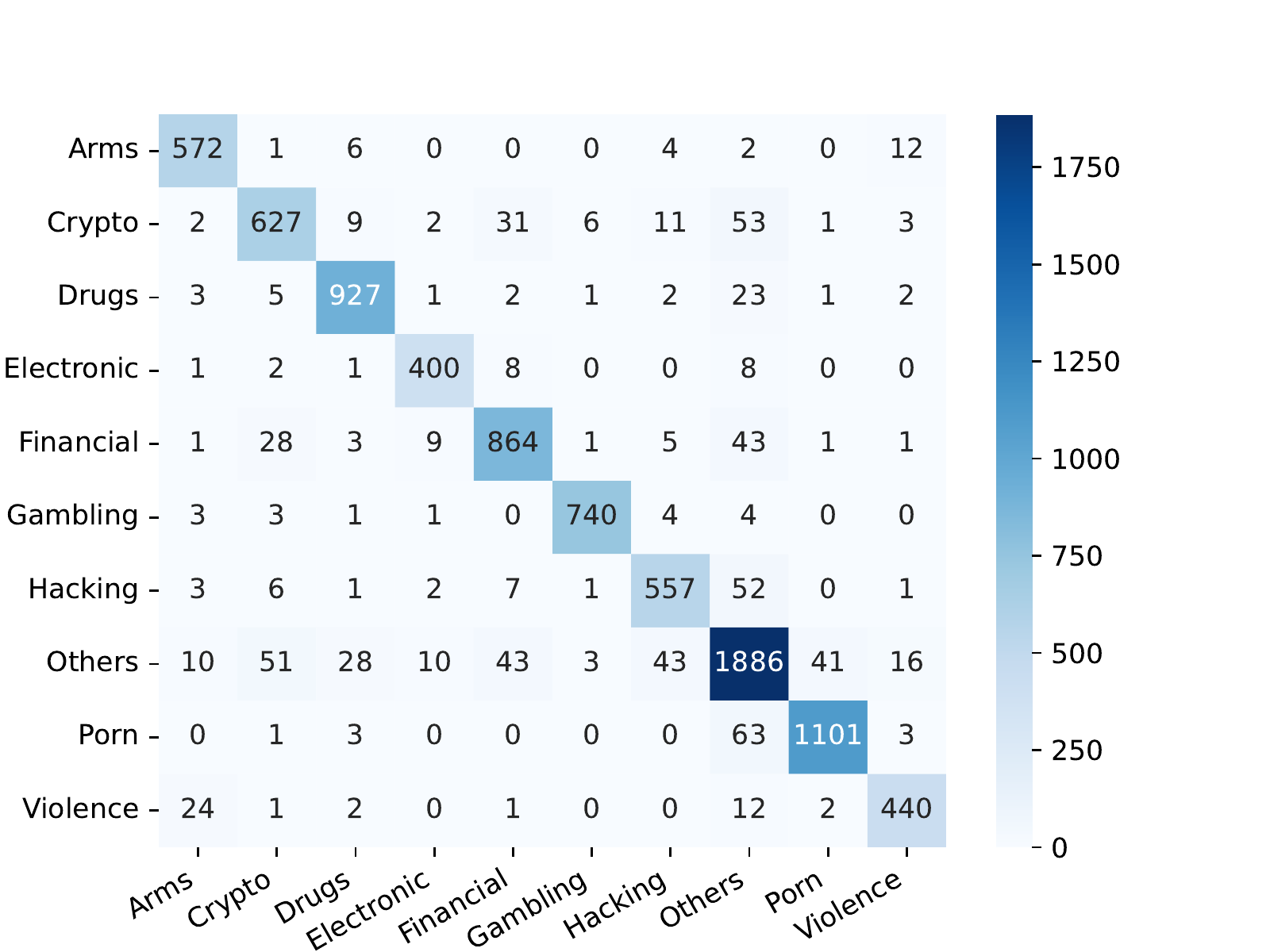}
         \caption{BERT\textsubscript{\textit{cased}}}
         \label{fig:cm-bert}
     \end{subfigure}
      \hspace{1em}
     \begin{subfigure}[b]{0.48\textwidth}
         \centering
         \includegraphics[width=\textwidth,trim={0 0 1.5cm 1.5cm},clip]{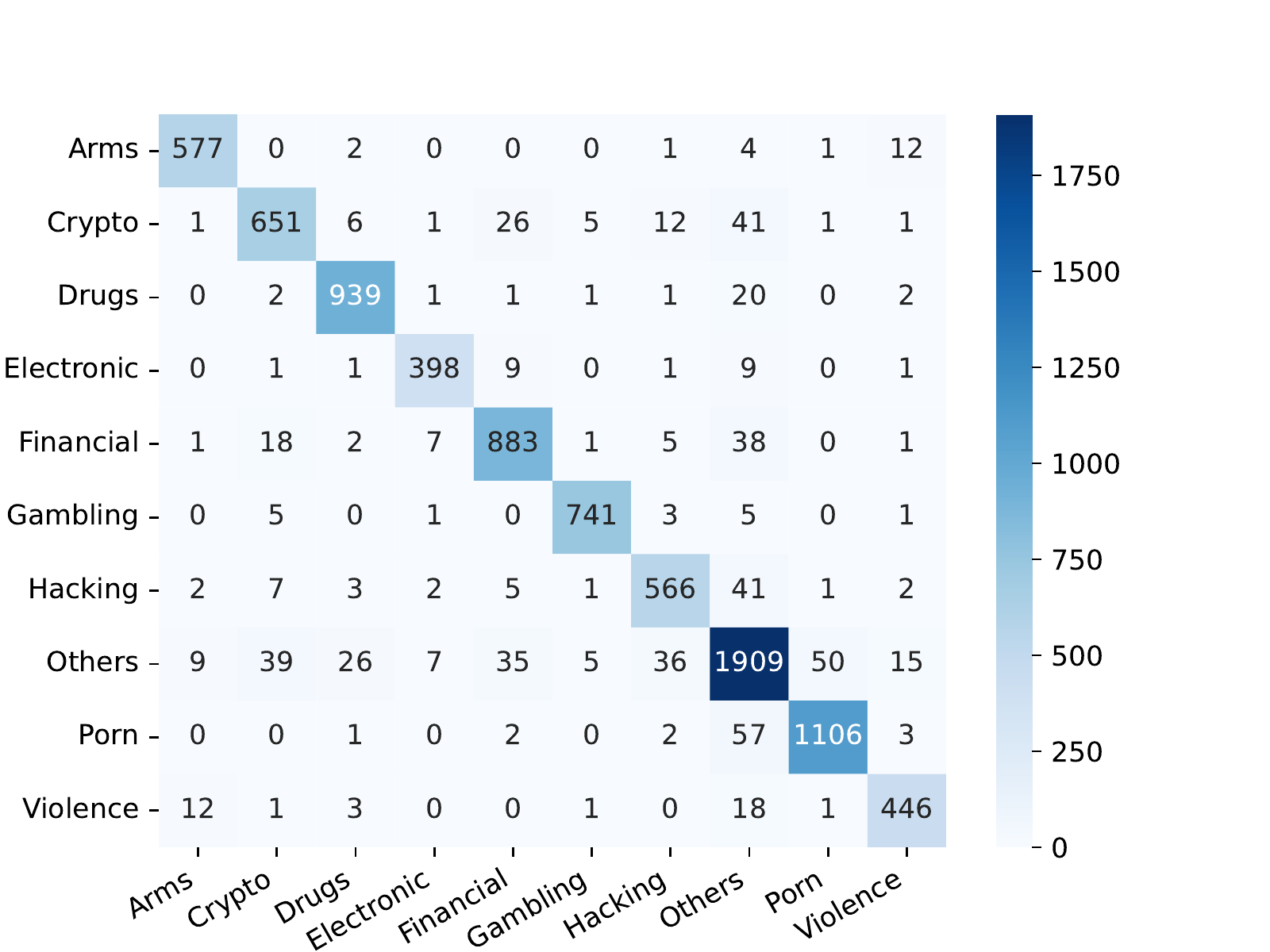}
         \caption{RoBERTa}
         \label{fig:cm-roberta}
     \end{subfigure}
     \vskip \baselineskip
     \begin{subfigure}[b]{0.48\textwidth}
         \centering
         \includegraphics[width=\textwidth,trim={0 0 1.5cm 1.5cm},clip]{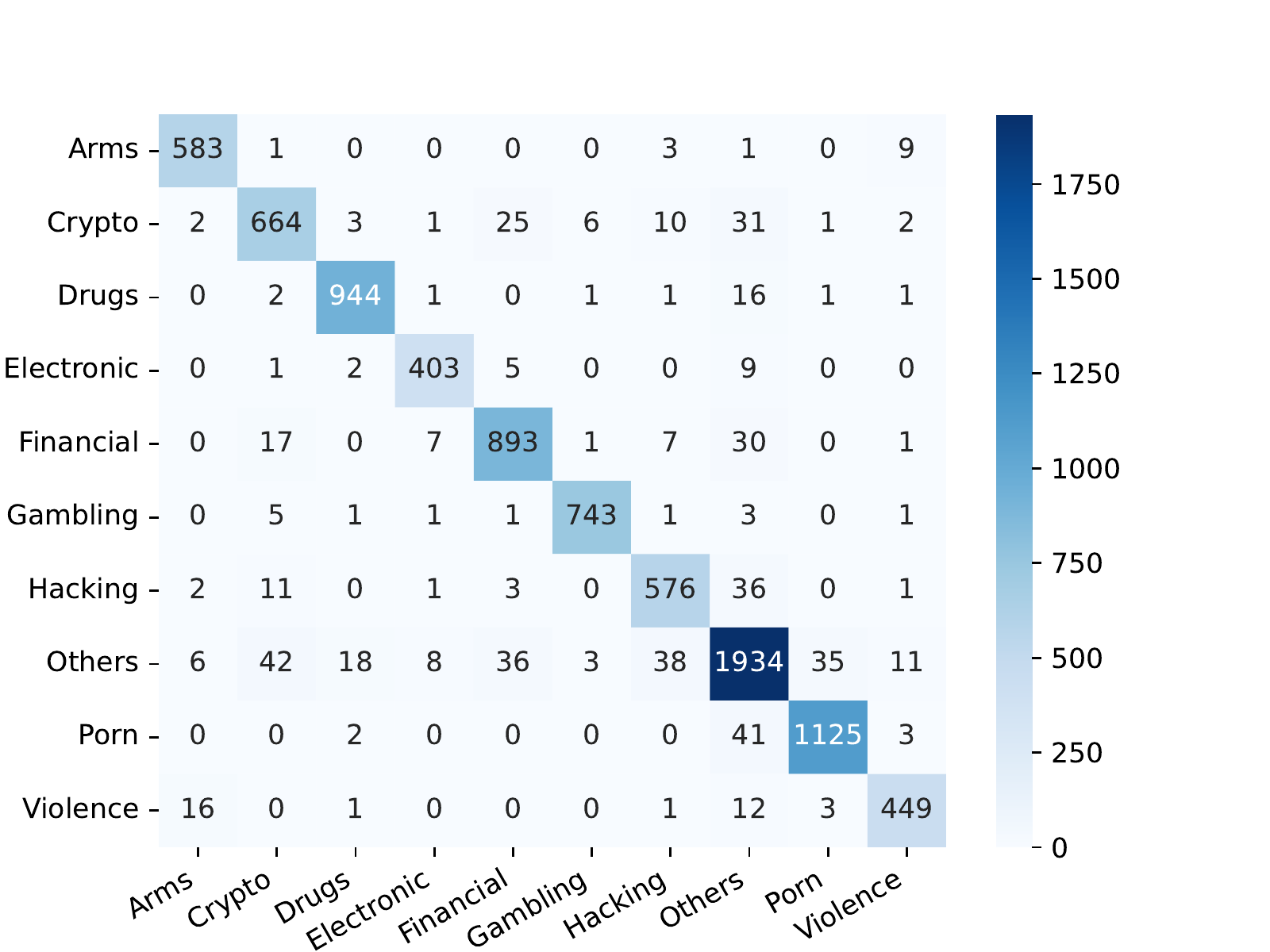}
         \caption{DarkBERT\textsubscript{\textit{raw}}}
         \label{fig:cm-darkbert-unproc}
     \end{subfigure}
      \hspace{1em}
     \begin{subfigure}[b]{0.48\textwidth}
         \centering
         \includegraphics[width=\textwidth,trim={0 0 1.5cm 1.5cm},clip]{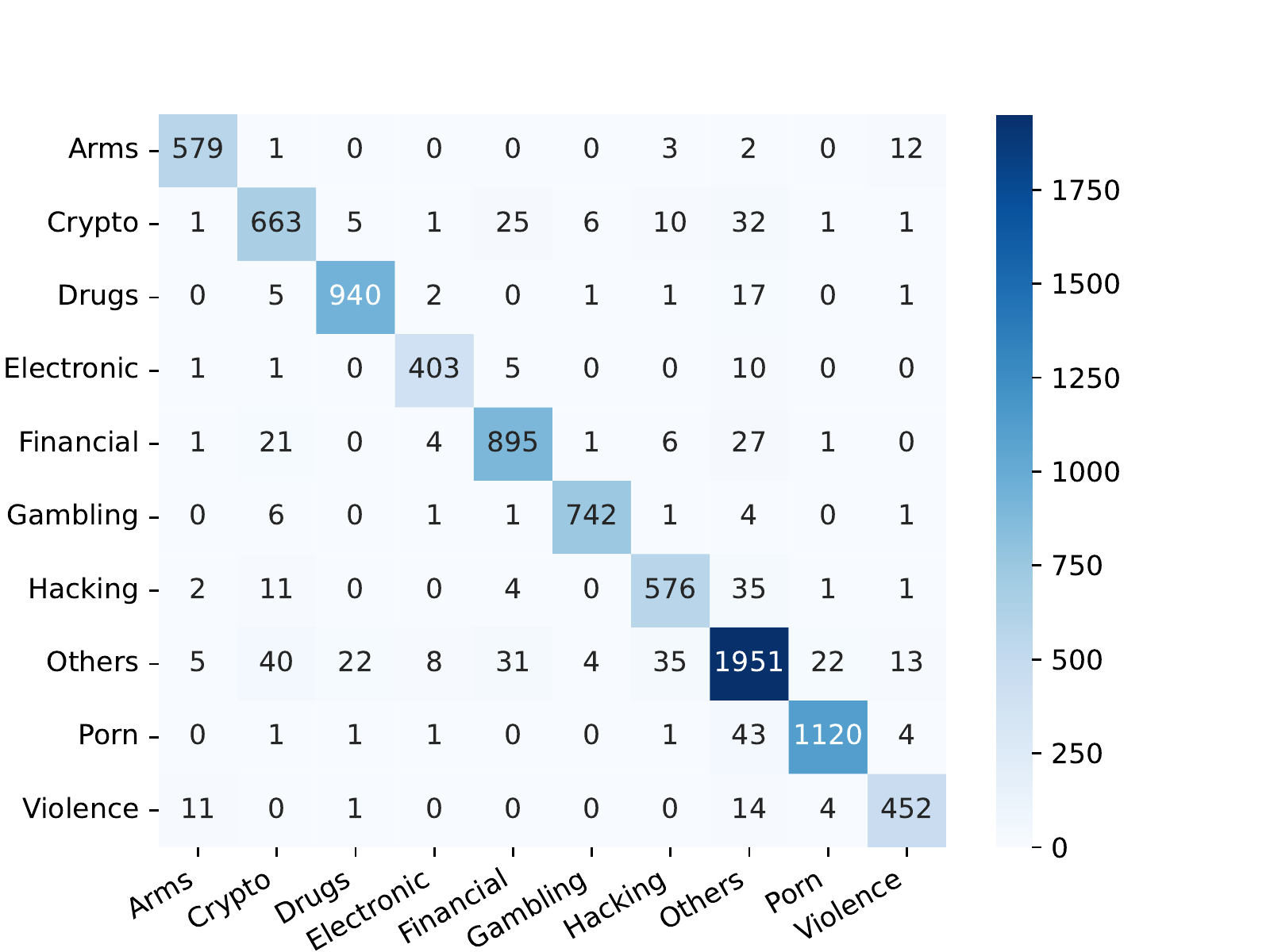}
         \caption{DarkBERT\textsubscript{\textit{preproc}}}
         \label{fig:cm-darkbert-preproc}
     \end{subfigure}
     \caption{Confusion matrices for selected language models evaluated on the CoDA\textsubscript{\textit{cased}} dataset}
    \label{fig:cm}
\end{figure*}

\end{document}